\RequirePackage{fix-cm}
\documentclass[twocolumn]{svjour3}          % twocolumn
\smartqed  % flush right qed marks, e.g. at end of proof
\usepackage{graphicx}
\usepackage{natbib}

\usepackage{times}
\usepackage{soul}
\usepackage{url}
\usepackage[utf8]{inputenc}
\usepackage[small]{caption}
\usepackage{graphicx}
\usepackage{amsmath}
\usepackage{amssymb}
\usepackage{booktabs}
\usepackage{algorithm}
\usepackage{algorithmic}
\usepackage[switch]{lineno}

\usepackage{tikz}
\usepackage{comment}
\usepackage{color}
\usepackage{array}
\allowdisplaybreaks[1]

\usepackage{diagbox}
\usepackage{multirow}
\usepackage{relsize}
\usepackage{bbm,dsfont}

\newcommand{\figref}[1]{Fig.~\ref{fig:#1}}
\newcommand{\tabref}[1]{Table~\ref{tbl:#1}}
\newcommand{\secref}[1]{Section~\ref{sec:#1}}

\DeclareMathOperator*{\argmin}{arg\,min}
\usepackage[pagebackref=true,breaklinks=true,colorlinks=True,bookmarks=false,linkcolor=blue,citecolor=blue]{hyperref}

\let\OLDthebibliography\thebibliography
\renewcommand\thebibliography[1]{
  \OLDthebibliography{#1}
  \setlength{\parskip}{0pt}
  \setlength{\itemsep}{0pt plus 0.3ex}
}

\newcommand{\figSota}{
\begin{figure}
 \centering
 \includegraphics[width=\linewidth]{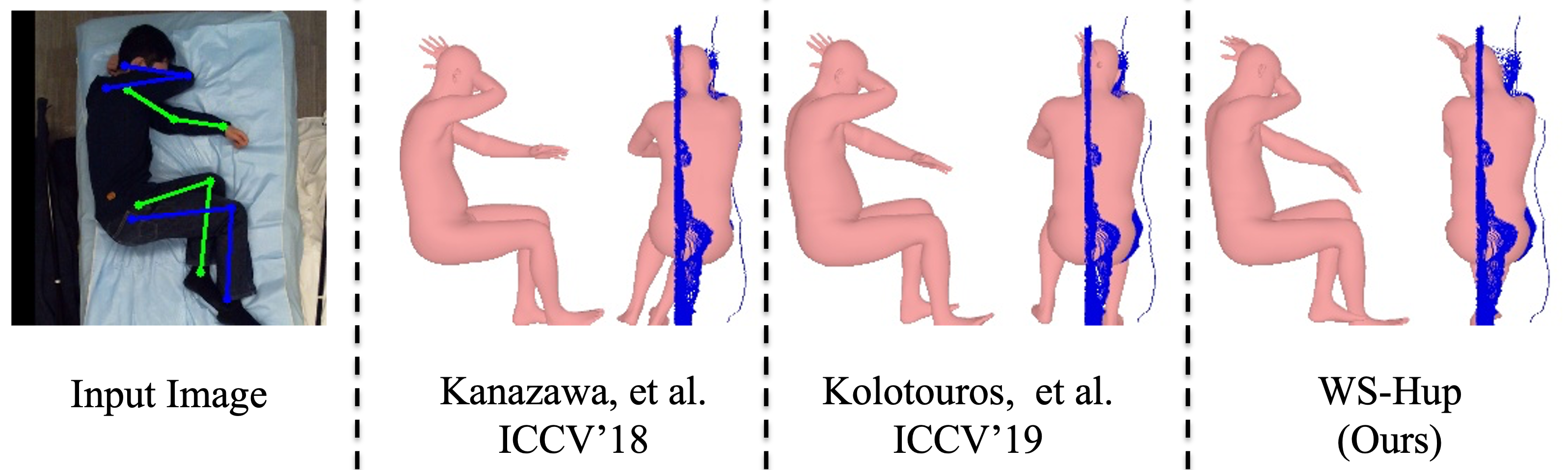}
 \caption{Predictions from SOTA 3D human pose and shape estimation models compared to our heuristic weakly-supervised 3D human pose estimation (HW-HuP) model, on an in-bed pose image taken from the SLP dataset \citep{liu2020simultaneously}. All models have been fine-tuned on SLP data. Viewed from the top, all three models perform well, but the side view shows that the HW-HuP estimation is more precise in a number of ways. Point clouds from the depth data are rendered in blue for reference.}
 \label{fig:sota}
\end{figure}
}

\newcommand{\figDepthProxy}{
\begin{figure}
 \centering
 \includegraphics[width=\linewidth]{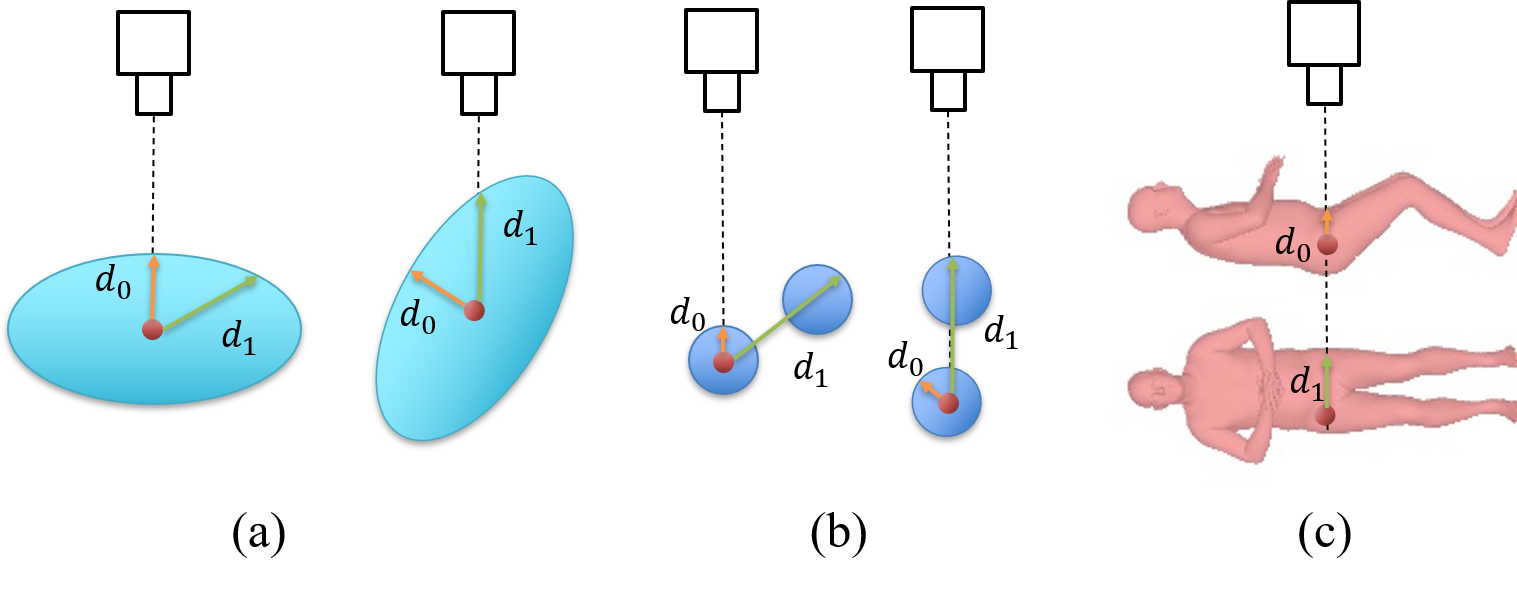}
 \caption{Depth proxy point bias from the true joint location in: (a) a single body with uneven shape, (b) double bodies with occlusion, and (c) right hip of a human body.}
 \label{fig:DepthProxy}
\end{figure}
}

\newcommand{\figHWS}{
\begin{figure*}
 \centering
 \includegraphics[width=0.7\textwidth]{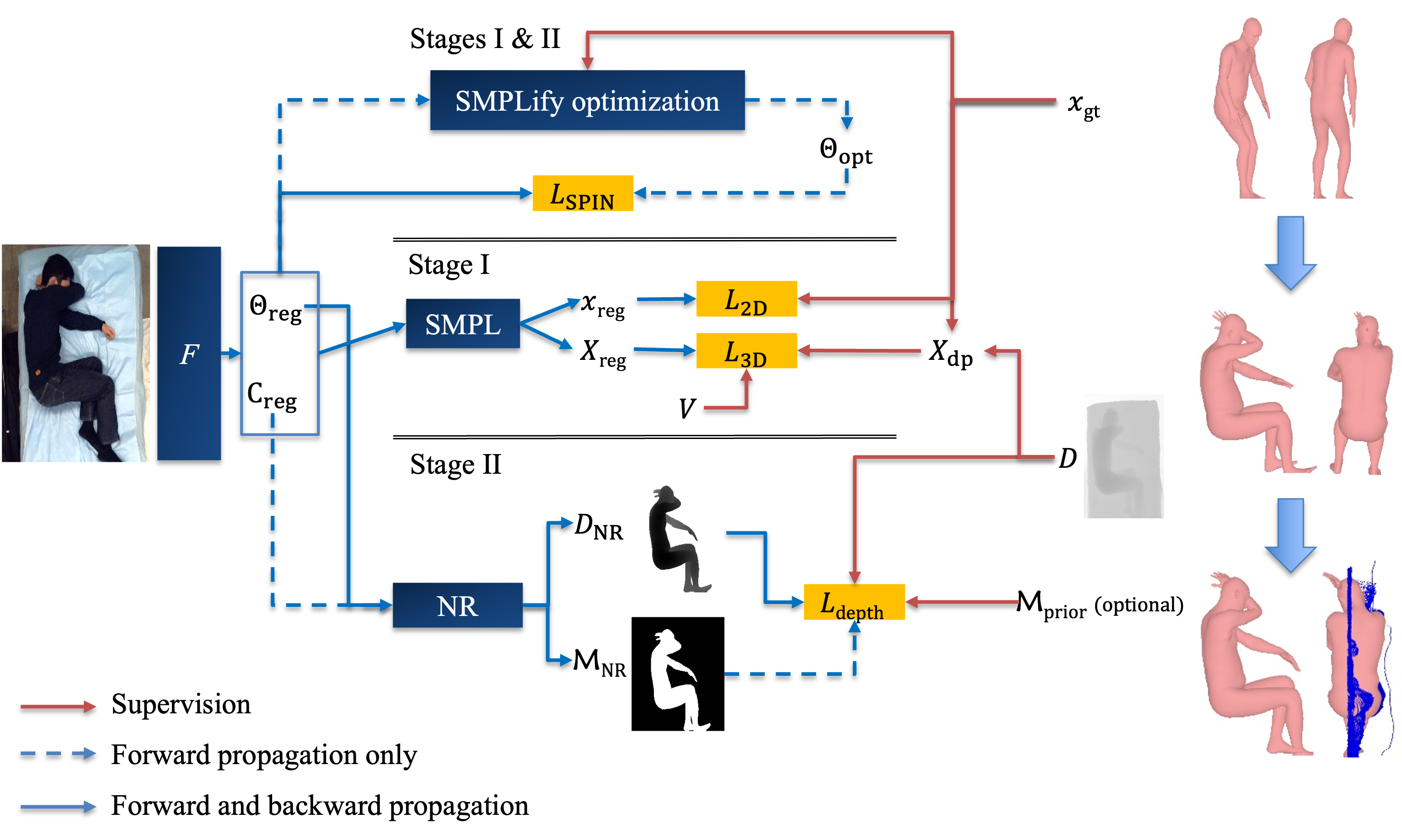}
 \caption{HW-HuP framework. In each step, HW-HuP trains the 3D pose regression function $F$ by supervising its pose predictions $[\Theta_\text{reg}, C_\text{reg}]$ with the outcome of the SMPLify 2D joint optimization, which incorporates heuristic prior information, as well as one of 3D based pipelines depending on the stage. In Stage I, HW-HuP uses coarse, depth based proxy 3D coordinates for the joints to supervise predicted joint positions, attenuated by visibility $V$ to ensure correct alignment of obscured limbs. After Stage I, the 2D joints and silhouette should be nearly aligned and the 3D pose acceptable. In Stage II, HW-HuP fine-tunes the 3D pose using the full depth map $D$ to supervise a depth map $D_\text{NR}$ generated by passing the predicted pose through a neural renderer NR, attenuated by a generated mask $\mathcal{M}_\text{NR}$ and an \emph{optional} mask prior $\mathcal{M}_\text{prior}$ if available.}
 \label{fig:HWS}
\end{figure*}
}

\newcommand{\figRGBuc}{
\begin{figure*}[ht]
 \centering
 \includegraphics[width=\linewidth]{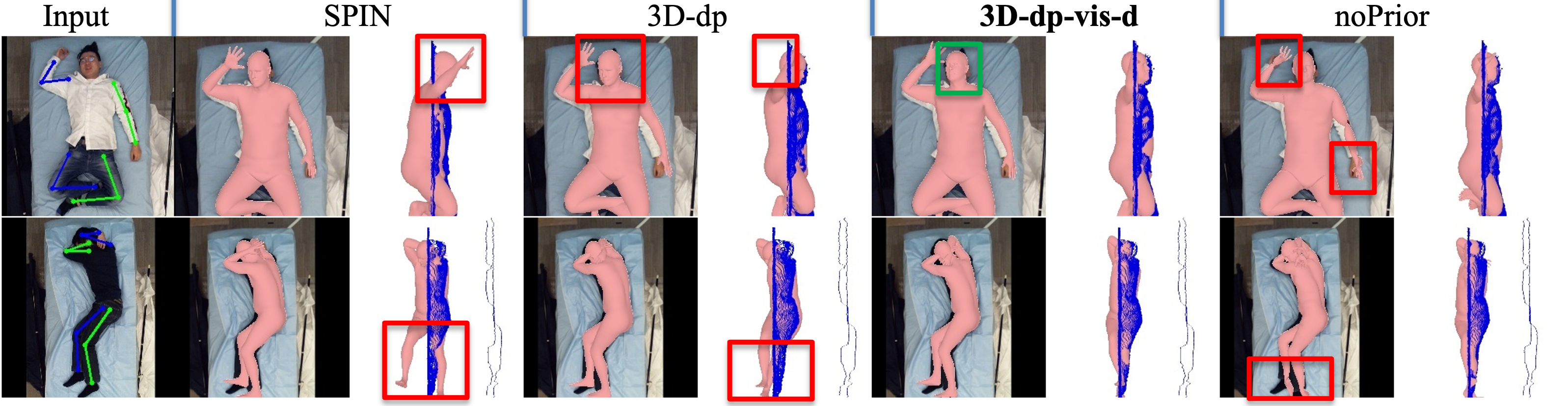}
 \caption{Qualitative 3D human pose and shape estimation on SLP dataset using ``Nocover'' RGB images. The first row features the supine posture, and the second the side lying posture. Point clouds are shown as blue dots. Unnatural or bad predictions are outlined in red, and the particularly successful head pose prediction from HW-HuP (i.e. 3D-dp-vis-d in our ablation study) is outlined by the green rectangle.}
 \label{fig:RGBuc}
\end{figure*}
}

\newcommand{\figAllC}{
\begin{figure}[ht]
 \centering
 \includegraphics[width=\linewidth]{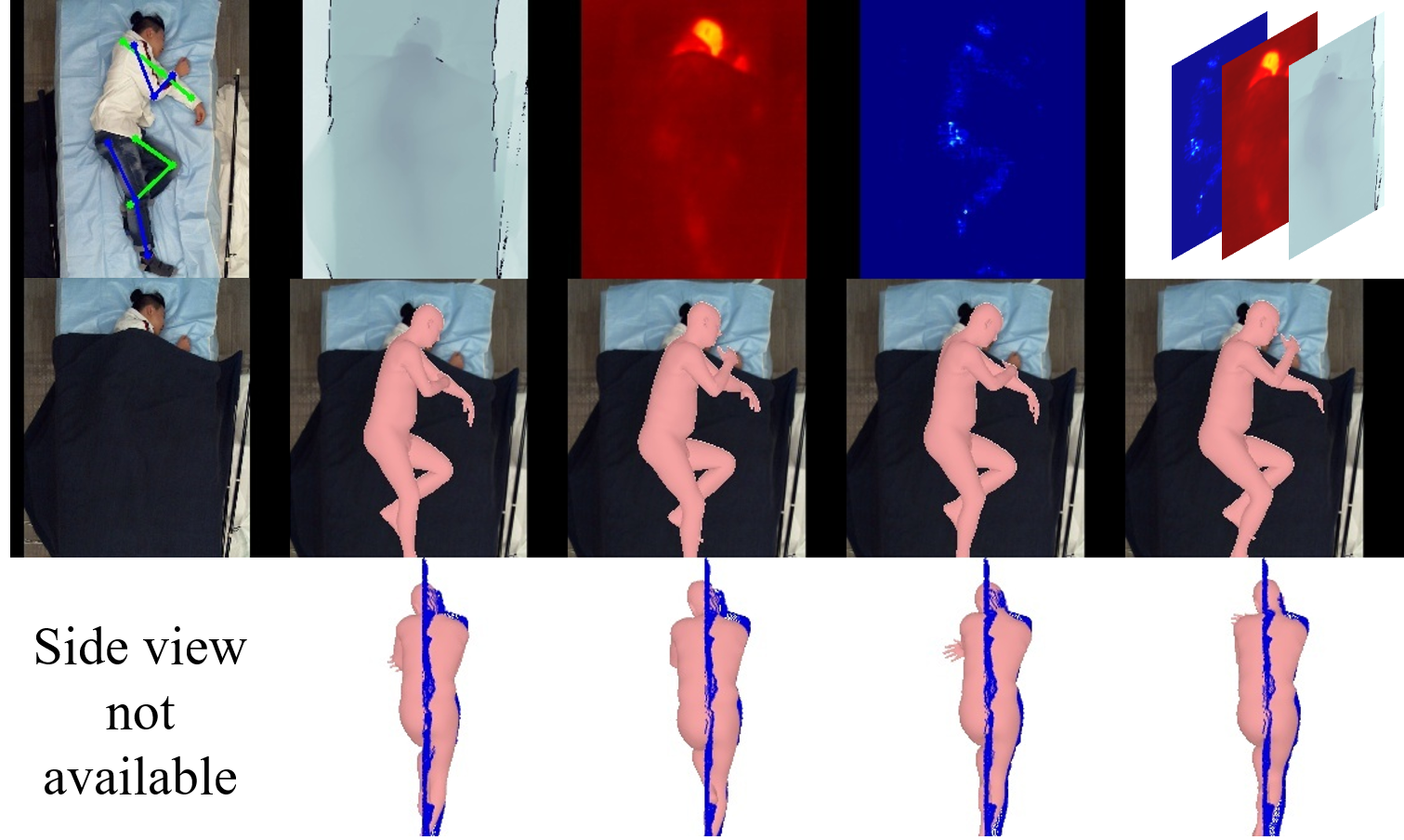}
 \caption{Qualitative 3D human pose and shape estimation results of our HW-HuP model applied to an example image from SLP dataset with heavy occlusion  (a thick blanket), under the respective input modalities of depth, long wavelength infrared (LWIR), pressure map (PM), and a combination of the three. First row shows the input modalities as well as a ``Nocover'' version of the RGB image as the reference. Second and third rows show the inference result of front view and side view, respectively.}
 \label{fig:allC}
\end{figure}
}

\newcommand{\fighm}{
\begin{figure}[ht]
 \centering
 \includegraphics[width=\linewidth]{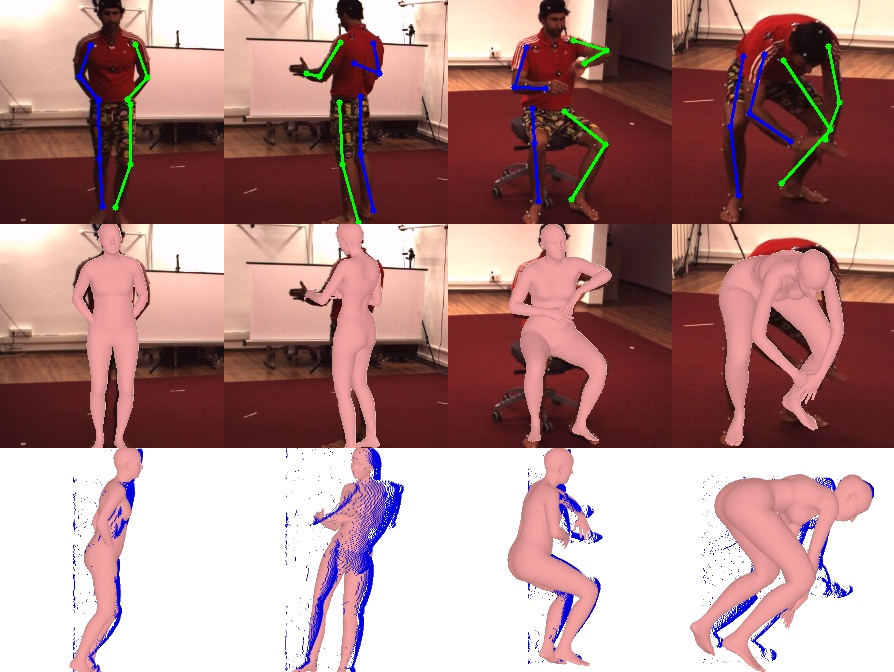}
 \caption{Qualitative 3D human pose and shape estimation results of our HW-HuP applied to the Human3.6m validation dataset.
 The first row shows the input RGB images, the second and 3r rows show HW-HuP pose predictions from the front and side view. Point clouds are rendered in blue dots for references. More visualization examples are displayed in the \textit{Supplementary Materials}.}
 \label{fig:hm}
\end{figure}
}

\newcommand{\figInfants}{
\begin{figure*}
 \centering
 \includegraphics[width=0.7\textwidth]{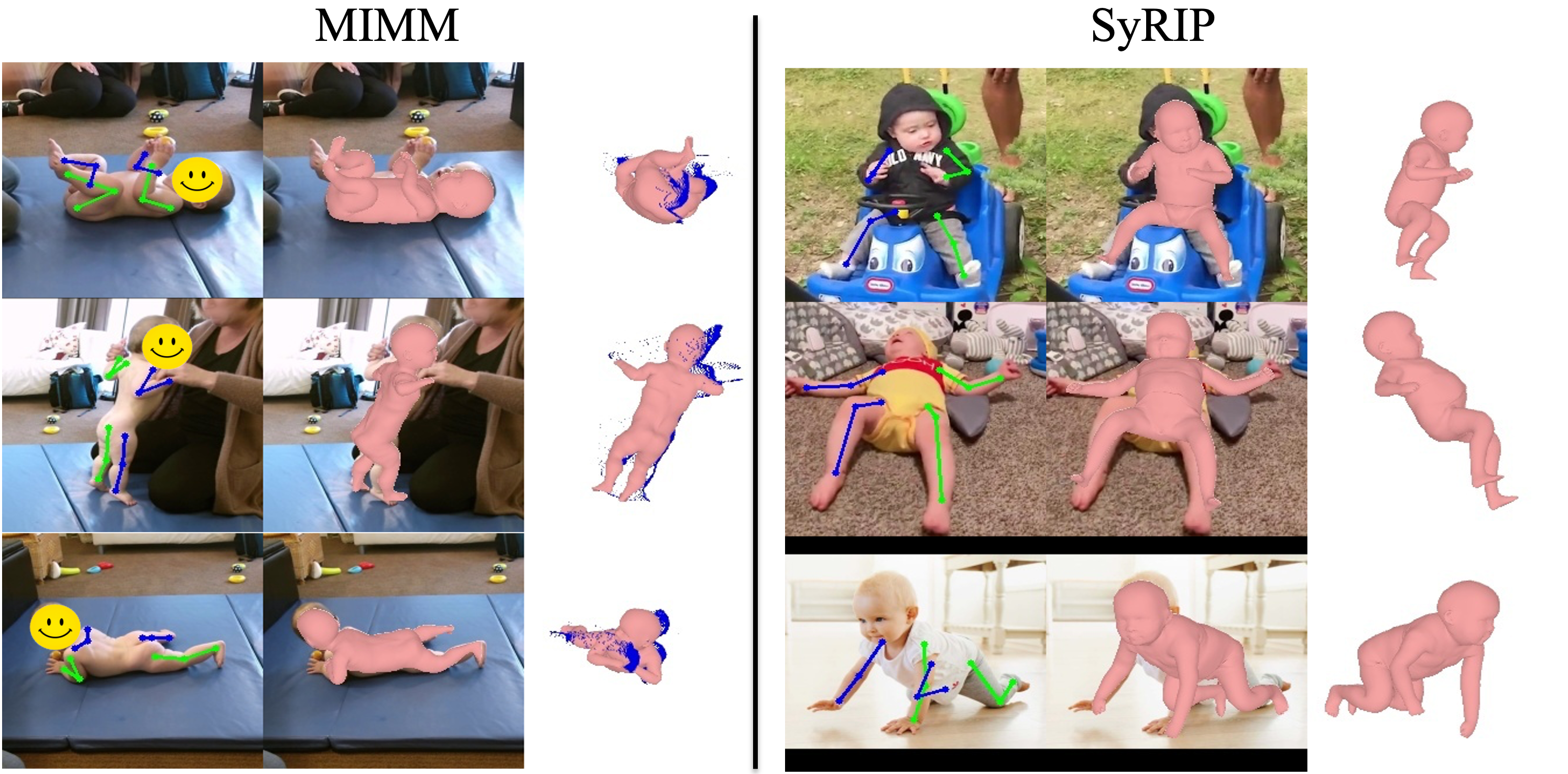}
 \caption{Qualitative 3D human pose and shape estimation results of our HW-HuP model applied on the MIMM and SyRIP infant datasets.
 The first column shows the input RGB image. The second and third columns visualize the HW-HuP results of front and side views, respectively. Point clouds (blue dots) are aligned with side view model for MIMM result.}
 \label{fig:infants}
\end{figure*}
}

\newcommand{\figSotaPre}{
\begin{figure}[ht]
 \centering
 \includegraphics[width=\linewidth]{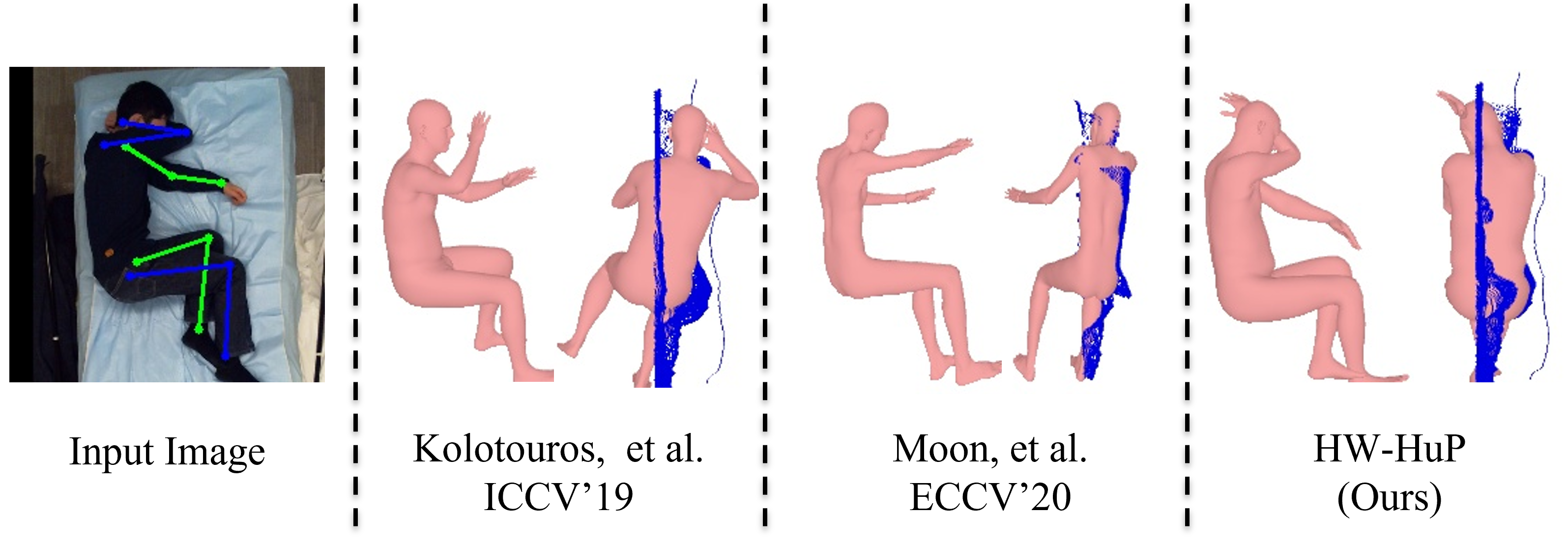}
 \caption{Visual inspection of the accuracy of the 3D human pose and shape estimation pre-trained models compared to our heuristic weakly-supervised 3D human pose estimation model (HW-HuP), when applied on an in-bed pose image taken from the SLP dataset \citep{liu2020simultaneously}. Point clouds from depth are rendered in blue for reference. Compared to \figref{sota} in the main paper, the SOTA models here are not fine-tuned on SLP data.}
 \label{fig:sotaPre}
\end{figure}
}

\newcommand{\figVisnet}{
\begin{figure}[t]
 \centering
 \includegraphics[width=\linewidth]{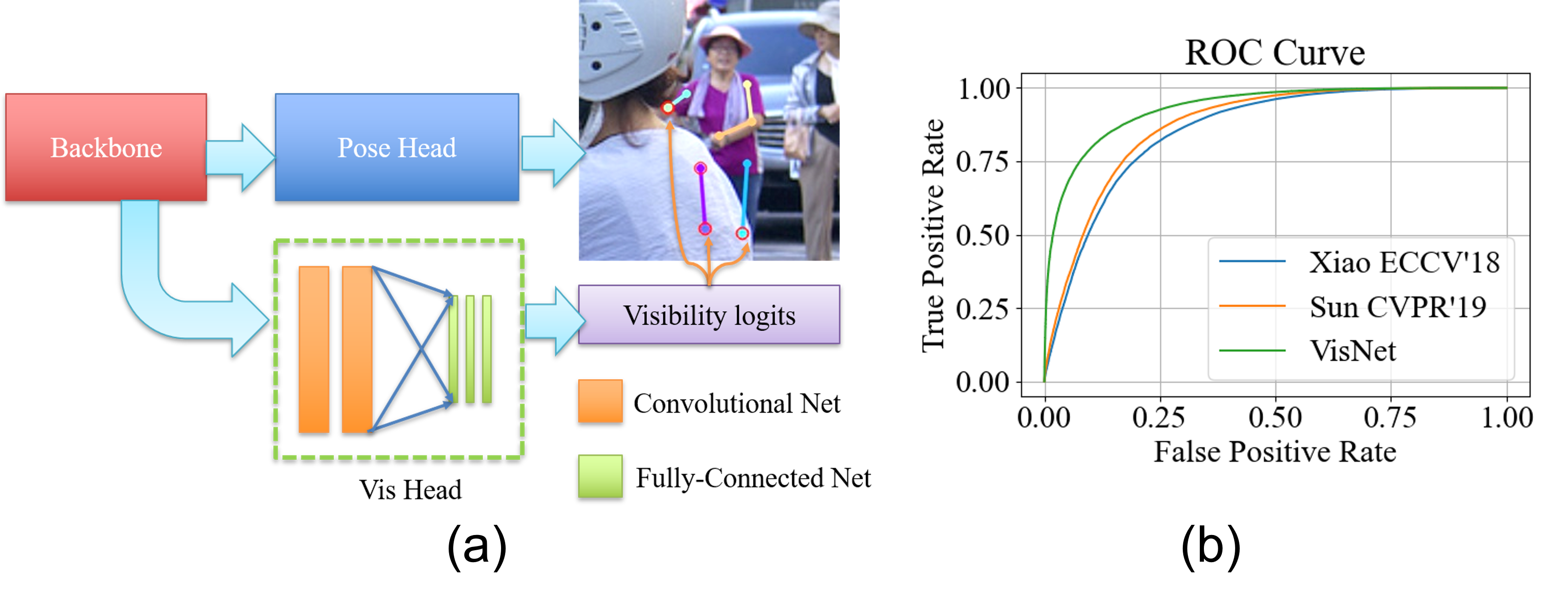}
 \caption{(a) VisNet model diagram, and (b) visibility detection ROC performance on COCO validation dataset in, comparing VisNet visibility scores to confidence scores from \citep{sun2019deep,xiao2018simple}.}
 \label{fig:visnet}
\end{figure}
}

\newcommand{\figVisnetGrid}{
\begin{figure}[ht]
 \centering
 \includegraphics[width=\linewidth]{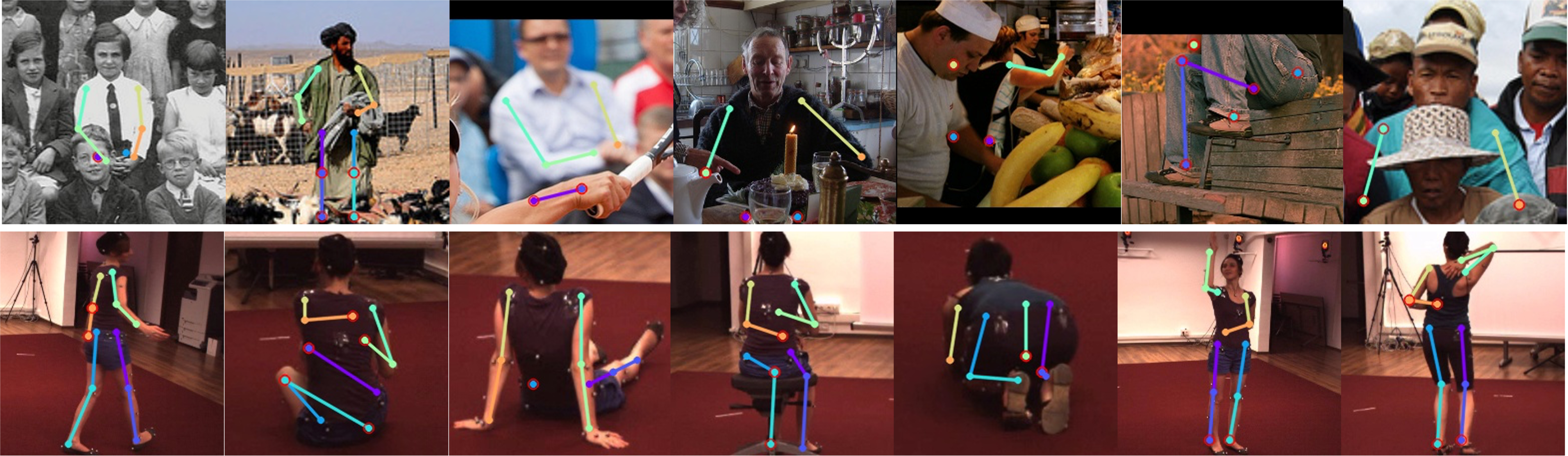}
 \caption{Qualitative results of VisNet on COCO (first row) and Human3.6M (second row). Detected occlusions are annotated with red circle.}
 \label{fig:visnetGrid}
\end{figure}
}
\newcommand{\figDpErr}{
\begin{figure*}[t]
 \centering
 \includegraphics[width=\linewidth]{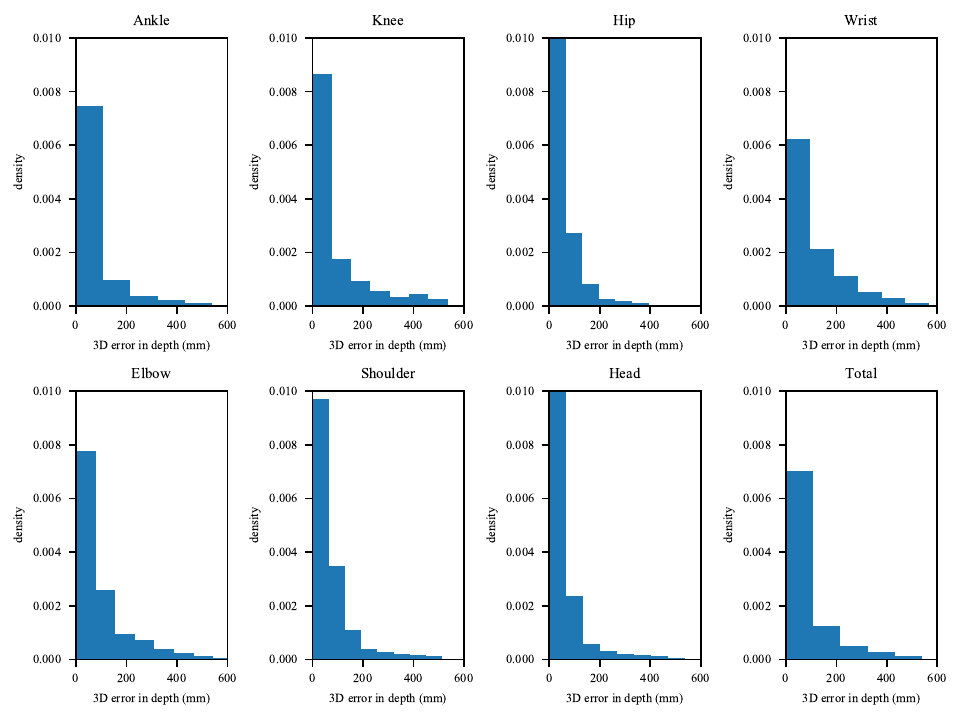}
 \caption{Shift error distribution between depth-based 3D proxy and real proxy by joints of Human3.6M training split.}
 \label{fig:dpErr}
\end{figure*}
}
\newcommand{\figAllCsup}{
\begin{figure*}[t]
 \centering
 \includegraphics[width=\linewidth]{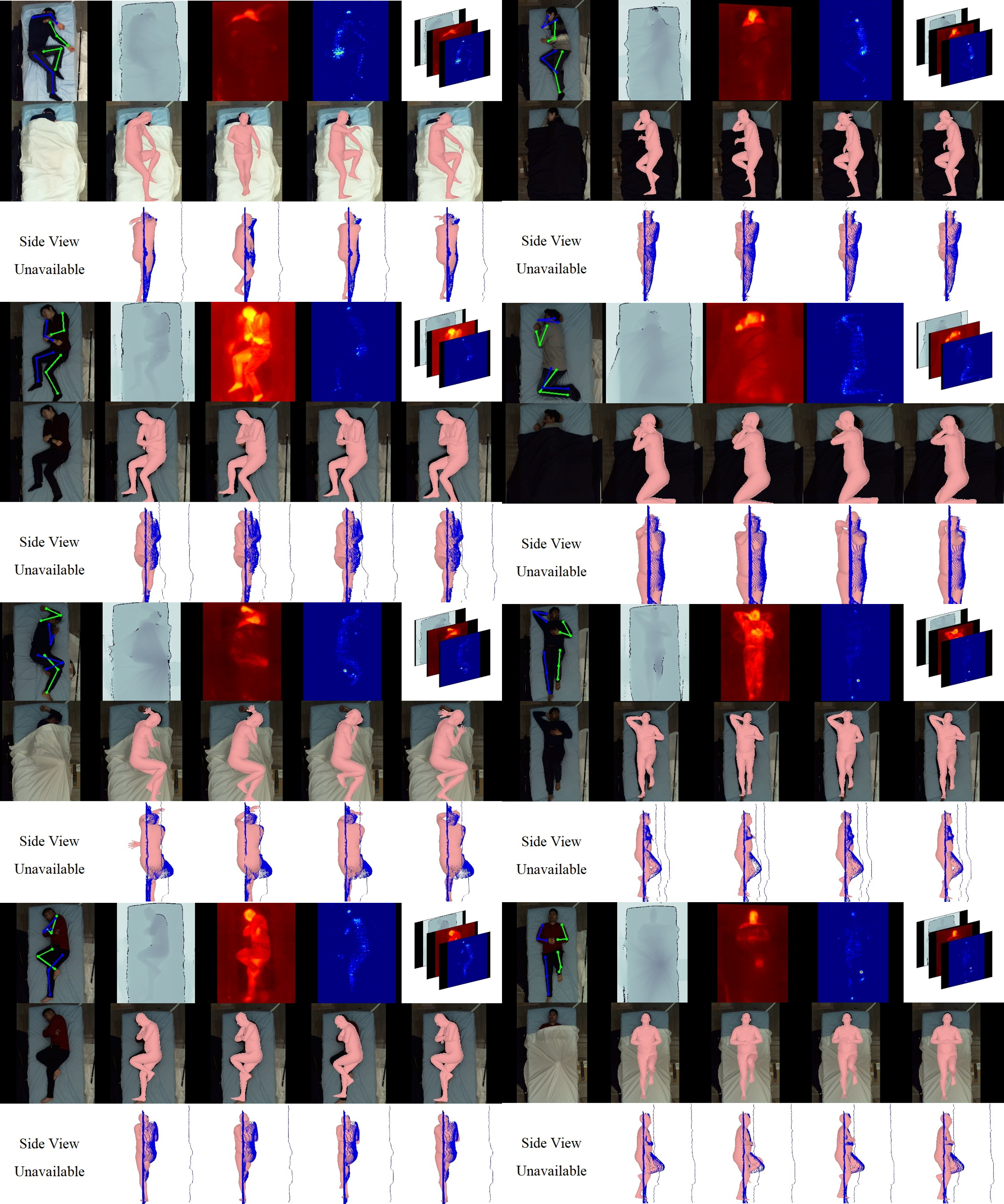}
 \caption{Qualitative 3D human pose and shape estimation results of our HW-HuP applied to the SLP dataset, when input 2D images consist of the depth, LWIR, PM, modalities, or their combination, respectively. The row shows the input modalities as well as a ``Nocover'' version of the RGB image for reference. The second and third rows show the inference result of front view and side view, respectively.}
 \label{fig:allCsup}
\end{figure*}
}

\newcommand{\fighmsup}{
\begin{figure*}[t]
 \centering
 \includegraphics[width=\linewidth]{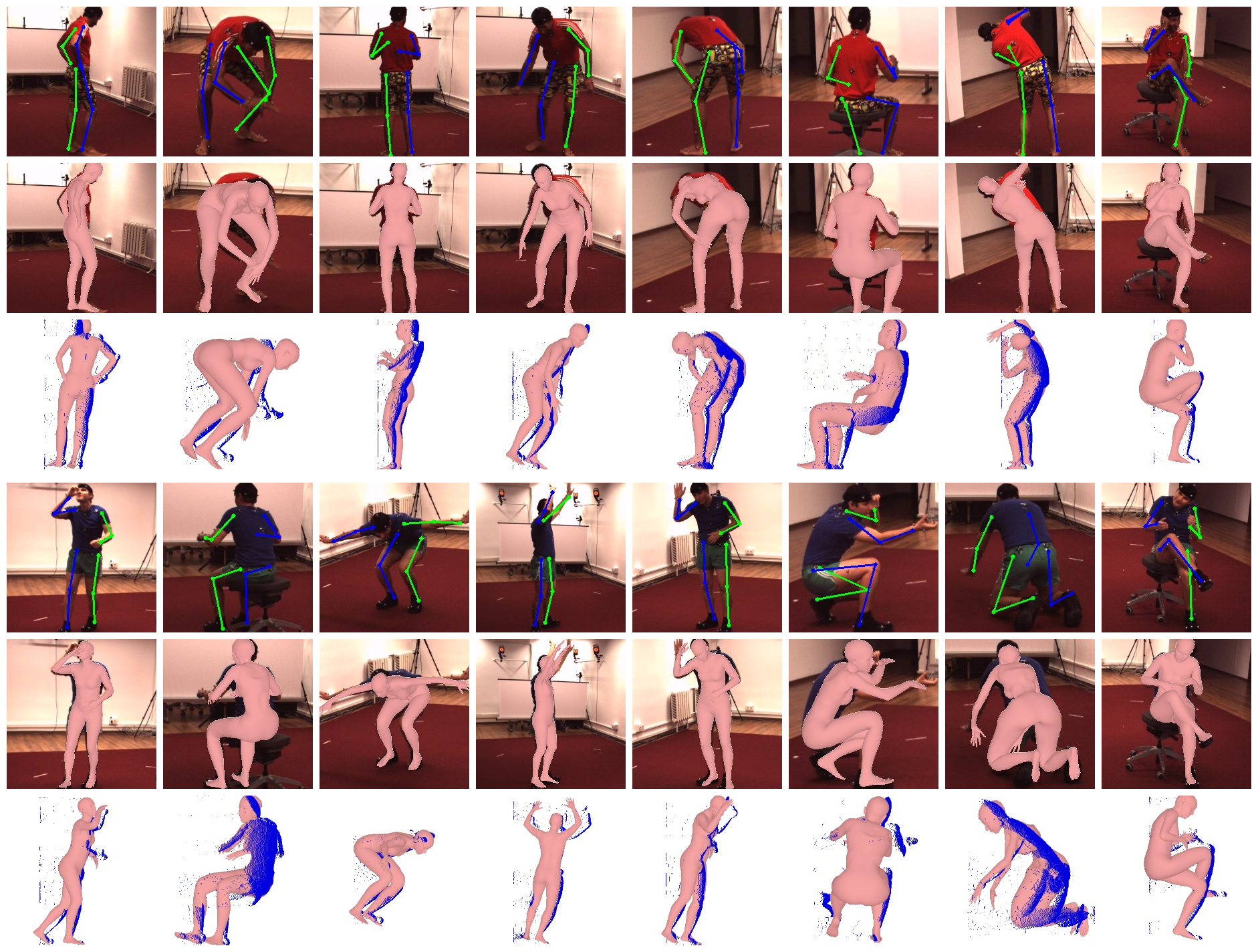}
 \caption{More qualitative 3D human pose and shape estimation results of our HW-HuP applied to the Human3.6m validation dataset. The first three rows show the predictions of frames for subject 9, and the last three rows for subject 11.}
 \label{fig:hmsup}
\end{figure*}
}

\newcommand{\tblAbla}{
\begin{table}
\caption{Aligned depth error (in mm) for 3D human pose and shape estimation predictions from SOTA models and HW-HuP configurations (for ablation), made on ``Nocover''  RGB images of the SLP dataset. Ablation studies defined in text.}
\begin{center}
\resizebox{\linewidth}{!}{
 \begin{tabular}{ccccc}   % 19 
 \hline
 \hline
  \multicolumn{5}{c}{SOTA Model}\\
  \hline
  SPIN* && SPIN && HMR\\
  80.10 && 68.38 && 63.43\\
  \hline 
  \hline 
  \multicolumn{5}{c}{HW-HuP \& Ablation Models}\\
  \hline 
  3D-dp & 3D-dp-vis & 3D-dp-vis-D & noPrior & 2D-D\\ 
 48.13 & 47.41 & \textbf{36.01} &38.54 & 39.32\\
 \hline
 \hline
 \end{tabular}
}
\label{tbl:abla}
\end{center}
\end{table}
}

\newcommand{\tblallcond}{
\begin{table}
\caption{Aligned depth error (in mm) of HW-HuP and SOTA 3D human pose and shape estimation on SLP dataset, under  ``Nocover'', ``Cover1'', ``Cover2'' and ``All Covers'' conditions; and under depth, LWIR, PM, and their ``Combined'' modalities.}
\begin{center}
\resizebox{\linewidth}{!}{
 \begin{tabular}{ l | *3r | r }   % 19 
 \hline
  \hline
 \textbf{Nocover}& Depth & LWIR & PM & Combined \\
  \hline
 SPIN*  & 108.78 & 89.41 & 102.27 & 96.13 \\
HMR  & 68.70 & 69.20 & 75.70 & 72.55 \\
SPIN  & 62.72 & 66.60 & 70.28 & 67.35 \\ 
\hline
\textbf{HW-HuP} & \textbf{33.87} & 38.45 & 41.48 & 37.62 \\
 \hline
  \hline
  \textbf{Cover1}& Depth & LWIR & PM & Combined \\
   \hline
  SPIN* & 105.82 & 92.65 & 101.01 & 102.27 \\
  HMR & 72.81 & 71.25 & 75.92 & 72.52\\
  SPIN  & 67.71 & 67.81 & 70.29 & 67.96\\
   \hline
  \textbf{HW-HuP}& \textbf{36.87} & 39.82 & 41.43 & 37.92 \\
 \hline
  \hline
\textbf{Cover2} & Depth & LWIR & PM & Combined\\
  \hline
 SPIN*  & 105.34 & 89.31 & 101.08 & 100.82 \\
HMR  & 73.06 & 70.69 & 76.10 & 73.11 \\
SPIN  & 68.38 & 66.58 & 70.40 & 68.15 \\
\hline
\textbf{HW-HuP} & \textbf{37.26} & 40.42 & 41.68 & 38.26 \\
\hline
 \hline
\text{All Covers} & Depth & LWIR & PM & Combined\\
\hline
SPIN* & 106.65 & 90.46 & 101.46 & 99.74 \\
HMR  & 71.52 & 70.38 & 75.91 & 72.73\\
SPIN & 66.27 & 67.00 & 70.32 & 67.82 \\
\textbf{HW-HuP} & \textbf{36.00} & 39.56 & 41.53 & 37.93\\ 
\hline
\hline
\end{tabular}
}
\label{tbl:allcond}
\end{center}
\end{table}
}
\newcommand{\tblHumanD}{
\begin{table}[h]
\caption{Comparison of HW-HuP with the SOTA based on the mean per joint position error (MPJPE) metric after  Procrustes analysis (PA) alignment with tested on Human3.6M dataset. HW-HuP performance is comparable to the SOTA with or without using ground truth 3D poses for training supervision. Approaches on the top part
require no image with 3D ground truth, while approaches on the
bottom part make use of 3D ground truth too. \textbf{Best PA MPJPE scores per data context in bold.}}
\begin{center}
\resizebox{\linewidth}{!}{
 \begin{tabular}{lrr}   % 19 
 \hline
\textbf{Method} & \textbf{Uses Paired 3D} & \textbf{PA MPJPE} \\
\hline
Lassner \citep{lassner2017unite} & No & 93.9 \\
Smplify \citep{bogo2016keep} & No  &  82.3 \\
Pavlakos et al. \citep{pavlakos2018learning} & No  & 75.9  \\
HMR (unpaired) \citep{kanazawa2018end} & No  & 66.5  \\
SPIN (unpaired) \citep{kolotouros2019learning} & No  & 62.0 \\
\hline
\textbf{HW-HuP (ours)}  & \textbf{No} & \textbf{50.4} \\
\hline
NBF \citep{omran2018neural} & Yes & 59.9 \\ 
HMR \citep{kanazawa2018end} & Yes & 56.8 \\ 
SPIN\citep{kolotouros2019learning} & Yes & \textbf{41.1} \\

DSD \citep{sun2019human} & Yes &       44.3  \\
I2L \citep{moon2020i2l} & Yes &       \textbf{41.1} \\
Pose2Mesh \citep{choi2020pose2mesh} & Yes & 46.3 \\
Song et al. \citep{song2020human} (w/2D kpts.) & Yes &    56.4 \\
VIBE \citep{kocabas2020vibe} (temporal) & Yes &    41.5 \\ 

\hline
\end{tabular}
}
\label{tbl:humanD}
\end{center}
\end{table}
}
\newcommand{\tblDPW}{
\begin{table}
\caption{Comparison of HW-HuP with the SOTA on the 3DPW dataset, based on the MPJPE metric. Note that HW-HuP does not use 3D ground truth training data, while the other models do. \textbf{Best PA MPJPE scores per data context in bold.}}
\begin{center}
\resizebox{\linewidth}{!}{
 \begin{tabular}{lrrr}   % 19 
 \hline
 Method & Uses 3D & MPJPE & PA-MPJPE \\
  \hline
HMR \citep{kanazawa2018end} & Yes  & - & 81.3 \\ 
GraphCMR \citep{kolotouros2019convolutional} & Yes  & - & 70.2  \\
SPIN \citep{kolotouros2019learning} & Yes & - & 59.2 \\
Pose2Mesh \citep{choi2020pose2mesh} & Yes &  89.2 & 58.9 \\
I2LMeshNet \citep{moon2020i2l} & Yes &  93.2 & 57.7 \\
VIBE \citep{kocabas2020vibe} & Yes & 82.0 & 51.9 \\
METRO \citep{lin2021end} & Yes & 77.1 & 47.9 \\
Mesh Graphormer \citep{lin2021mesh} & Yes & \textbf{74.7} &  \textbf{45.6} \\
\hline
HW-HuP & No & - & \textbf{66.1} \\
\hline
\end{tabular}}
\label{tbl:DPW}
\end{center}
\end{table}
}
\newcommand{\tblpck}{
\begin{table*}[ht]
\caption{PCK@0.2 of HW-HuP and SOTA 3D joint predictions projected to 2D, compared to 2D ground truth, under ``Nocover,'' ``Cover1,'' ``Cover2'' and ``All Covers'' conditions and depth, LWIR, PM modalities in the SLP dataset.}
\begin{center}
\resizebox{\linewidth}{!}{
 \begin{tabular}{ c | *3c | c || c | *3c | c  }   % 19 
 \hline
 Nocover& Depth & LWIR & PM & Combined & Cover1& Depth & LWIR & PM & Combined \\
  \hline
 SPIN* \citep{kolotouros2019learning}   & 34.43 & 42.85 & 15.49 & 45.90 &SPIN *\citep{kolotouros2019learning} & 19.63 & 17.21 & 15.82 & 24.58 \\
HMR \citep{kanazawa2018end} & 96.51 & 95.57 & 90.59 & 95.46 &HMR \citep{kanazawa2018end} & 90.76 & 91.93 & 90.71 & 94.14\\
SPIN \citep{kolotouros2019learning} & 96.23 & 95.57 & 90.56 & 95.45  &SPIN \citep{kolotouros2019learning} & 90.37 & 92.21 & 90.51 & 94.17\\
\hline
\textbf{HW-HuP} & \textbf{96.64} & 95.22 & 91.19 &95.49 &\textbf{HW-HuP}& 91.70 & 91.90 & 91.20 & \textbf{94.21} \\
 \hline
 \hline
Cover2 & Depth & LWIR & PM & Combined & All Covers & Depth & LWIR & PM & Combined\\
  \hline
 SPIN* \citep{kolotouros2019learning}  & 21.27 & 17.61 & 15.79 & 25.14 & SPIN* \citep{kolotouros2019learning} & 25.11 & 25.89 & 15.70 & 31.87 \\
HMR \citep{kanazawa2018end} & 91.17 & 91.53 & 90.80 & 93.38 &HMR \citep{kanazawa2018end} & 92.81 & 93.01 & 90.70 & 94.33\\
SPIN \citep{kolotouros2019learning} & 90.83 & 91.94 & 90.34 & 93.33  &SPIN \citep{kolotouros2019learning}& 92.48 & 93.24 & 90.47 & 94.32 \\
\hline
\textbf{HW-HuP} & 91.68 & 91.76 & 90.79 & \textbf{93.77} &\textbf{HW-HuP} & 93.34 & 92.96 & 91.06 & \textbf{94.49}\\ 
\hline
\end{tabular}}
\label{tbl:pck}
\end{center}
\end{table*}
}

\newcommand{\tblHumanCall}{
\begin{table*}
\caption{Comparison of HW-HuP with the SOTA based on the mean per joint position error (MPJPE) metric tested on Human3.6M dataset using Protocol\#2, including some results with 
rigid Procrustes analysis (PA) alignment \citep{gower1975generalized}. HW-HuP performance is comparable to the SOTA despite not using ground truth 3D poses for training supervision. \textbf{Best MPJPE scores}, \textit{best PA MPJPE scores}, and models (*) with further temporal constraints indicated.}
\begin{center}
\small
\resizebox{\linewidth}{!}{
 \begin{tabular}{c  c   c    c     c  c   c  c   c    c     c   c   c  c   c    c     c  }   % 19 
 \hline
Methods  & Dir. & Dis. & Eat & Gre. & Phon. & Pose & Pur. & Sit & SitD. & Smo. & Phot. & Wait & Walk & WalkD. & WalkT. & Avg \\
\hline
Zhou et al. \citep{zhou2016sparse} &
99.7 & 95.8 & 87.9 & 116.8 & 108.3 & 107.3 & 93.5 & 95.3 & 109.1 & 137.5 & 106.0 & 102.2 & 106.5 & 110.4 & 115.2 & 106.7 \\
SMPLify et al. \citep{bogo2016keep} & 
62.0 & 60.2 & 67.8 & 76.5 & 92.1 & 77.0 & 73.0 & 75.3 & 100.3 & 137.3 & 83.4 & 77.3 & 79.7 & 86.8 & 81.7 & 82.3  \\ 
Zhou et al. \citep{zhou2018monocap} & 68.7 & 74.8 & 67.8 & 76.4 & 76.3 & 84.0 & 70.2 & 88.0 & 113.8 & 78.0 & 98.4 & 90.1 & 62.6 & 75.1 & 73.6 & 79.9 \\
Martinez et al. \citep{martinez2017simple} & 51.8 & 56.2 & 58.1 & 59.0 & 69.5 & 55.2 & 58.1 & 74.0 & 94.6 & 62.3 & 78.4 & 59.1 & 49.5 & 65.1 & 52.4 & 62.9 \\
Sun et al. \citep{fang2018learning} & 47.5 & 47.7 & 49.5 & 50.2 & 51.4 & 43.8 & 46.4 & 58.9 & 65.7 & 49.4 & 55.8 & 47.8 & 38.9 & 49.0 & 43.8 & 49.6 \\
Moon et al. \citep{moon2019camera} & 50.5 & 55.7 & 50.1 & 51.7 & 53.9 & 46.8 & 50.0 & 61.9 & 68.0 & 52.5 & 55.9 & 49.9 & 41.8 & 56.1 & 46.9 & 53.3 \\
Iskakov et al. (monocular) \citep{iskakov2019learnable} &  41.9 & 49.2 & 46.9 & 47.6 & 50.7 & 57.9 & 41.2 & 50.9 & 57.3 & 74.9 & 48.6 & 44.3 & 41.3 & 52.8 & 42.7 & 49.9 \\
Pavllo et al. * \citep{pavllo20193d} & 45.2 & 46.7 & 43.3 & 45.6 & 48.1 & 55.1 & 44.6 & 44.3 & 57.3 & 65.8 & 47.1 & 44.0 & 49.0 & 32.8 & 33.9 & 46.8  \\
Cheng et al. * \citep{cheng2019occlusion} & 38.3 & 41.3 & 46.1 & 40.1 & 41.6 & 51.9 & 41.8 & 40.9 & 51.5 & 58.4 & 42.2 & 44.6 & 41.7 & 33.7 & 30.1 & 42.9 \\
Cheng et al. * \citep{cheng20203d} & 36.2 & 38.1 & 42.7 & 35.9 & 38.2 & 45.7 & 36.8 & 42.0 & 45.9 & 51.3 & 41.8 & 41.5 & 43.8 & 33.1 & 28.6 & \textbf{40.1} \\
Reddy et al. * \citep{reddy2021tessetrack} & 38.4 & 46.2 & 44.3 & 43.2 & 44.8 & 48.3 & 52.9 & 36.7 & 45.3 & 54.5 & 63.4 & 44.4 & 41.9 & 46.2 & 39.9 & 44.6 \\
\hline
SPIN et al. \citep{kolotouros2019learning}&  60.6 & 61.0 & 57.9 & 64.5 & 67.1 & 59.9 & 57.7 & 80.1 & 91.3 & 63.2 & 65.8 & 60.2 & 53.2 & 61.2 & 60.4 & 65.7 \\
SPIN et al. PA  & 39.5 & 42.1 & 41.1 & 44.8 & 46.6 & 38.6 & 38.8 & 60.8 & 68.5 & 45.2 & 46.2 & 41.4 & 36.0 & 45.0 & 40.9 & \textit{44.2} \\
\hline
\hline
HW-HuP &  95.0 & 103.9 & 99.8 & 101.0 & 106.3 & 94.2 & 110.1 & 121.4 & 140.1 & 103.3 & 112.8 & 99.5 & 92.8 & 107.2 & 101.9 & 104.1 \\
HW-HuP PA &  47.0 & 49.5 & 49.2 & 51.2 & 52.1 & 45.0 & 48.4 & 64.7 & 75.4 & 49.9 & 54.3 & 47.8 & 42.6 & 53.3 & 48.8 & 50.4 \\
\hline
\end{tabular}
}
\label{tbl:humanCall}
\end{center}
\end{table*}
}

\newcommand{\supp}{
\newcommand{\beginsupplement}{%
        \setcounter{table}{0}
        \setcounter{equation}{0}
        \setcounter{section}{0}
        \renewcommand{\thetable}{S\arabic{table}}%
        \setcounter{figure}{0}
        \renewcommand{\thefigure}{S\arabic{figure}}%
}
\renewcommand\thesection{\Alph{section}}
\beginsupplement
\clearpage

\section{Supplementary Materials}
Our source code is included as part of our supplementary submission. Furthermore, we discuss here: (1) a qualitative comparison between the HW-HuP and pre-trained SOTA 3D pose estimation models, (2) the design and evaluation of the VisNet for joint visibility estimation, (3) the results of projected 2D pose estimation on the SLP dataset, (4) concepts pertaining to selective pose prior transfer, (5) the error distribution of the depth-based 3D proxy, (6) a detailed comparison with the SOTA on Human3.6M dataset, and finally, (7) additional qualitative results. 

\subsection{Qualitative Performance of SOTA Pre-trained Model}
Given that many SOTA models already show satisfactory performance on large-scale 3D pose benchmarks such as Human3.6M \cite{ionescu2013human3} and 3DPW \cite{kolotouros2019convolutional}, one might wonder whether the pretrained models are already sufficient for real life applications, even without domain-specific fine-tuning. We examined this question qualitatively for in-bed images from the SLP dataset, where poses appear simpler. A illustrative example of the pretrained model performance is given in \figref{sotaPre}. Contrast this with the better performance of the SOTA models after fine-tuning in \figref{sota}---although even with fine-tuning, HW-HuP achieves superior results.
\figSotaPre

\subsection{VisNet for Joint Visibility Estimation}
The main purpose of VisNet is to determine the visibility of joints in an image of a body. This issue has not featured extensively in existing pose estimation models since many such 3D models are trained on MoCap data, where visibility is not an issue, and otherwise pose estimation models are assumed to be capable of inferring the position of occluded joints regardless. However, as discussed in \secref{PF}, such occlusions can yield large biases in the depth-based proxy 3D joint coordinates used by HW-HuP to supervise its pose regression, precipitating the need for VisNet to filter out unreliable joints. In our design, the VisNet head is based on a ResNet \cite{he2016deep} backbone. It includes two convolution layers with $1 \times 1$ kernels and 256 and 32 channels respectively,  followed by three fully connected layers with 256, 64, and 17 channels respectively. Each layer is followed by a batch normalization and a rectified linear unit (ReLU). To enhance its semantic understanding of the specific joint for visibility detection, we add the pose head \cite{xiao2018simple} on top of the backbone for joint training. VisNet design is shown in \figref{visnet}(a). In our implementation, VisNet is trained with an Adam optimizer with learning rate of 0.001 with a total epoch of 80 trained on COCO dataset \cite{lin2014microsoft}, which contains visibility annotations.
As a simple baseline quantification of VisNet performance, we compare its predicted per joint visibility scores against the per joint prediction confidence scores obtained from two SOTA human pose estimation models \cite{xiao2018simple,sun2019deep}, with receiver operating characteristic curves shown in \figref{visnet}(b). With the area under the curve (AUC) of the ROC curve serving as our metric, VisNet scores 93.3\%, higher than 85.6\% from \cite{xiao2018simple} and 87.5\% from \cite{sun2019deep}. Qualitative results of VisNet on COCO and Human3.6M \cite{ionescu2013human3} are shown in \figref{visnetGrid}. We observe that VisNet can more easily detect occlusions when the blocking and blocked objects have distinct appearances from each other. We also observe some common failures cases: (1) Visible joints can we wrongly classified as occluded when they are close to the boundary or to other objects, as with the image in \figref{visnetGrid} row 2, column 6. Presumably, sharp changes such as boundaries between objects or image boundaries are taken as clues of occlusion. (2) If the image only features partial views of the limbs,  as in \figref{visnetGrid} row 1, column 6, the joints may be wrongly classified as occluded.
(3) If one of the joints is occluded, the neighboring joints on the same limb are sometimes also deemed to be occluded, such as \figref{visnetGrid} (r2 c7).  

\figVisnet 
\figVisnetGrid

\subsection{Projected 2D Pose Estimation Results on SLP}
\tblpck
In \tabref{allcond} of the main paper, we demonstrated the superior performance of HW-HuP compared to SOTA on the SLP dataset by comparing their 3D pose predictions with the aligned depth error metric. Here, we offer another perspective on the same comparison, by projecting the 3D joint predictions of each model down to 2D and measuring the resulting percentage of correct keypoints (PCK) \cite{6380498} relative to ground truth 2D annotations. \tabref{pck} plots the results, with PCK correctness acceptance threshold of 20\% of the body torso size (PCK@0.2). Since all models incorporate components of successful 2D pose estimators, the fine-tuned predictions are all quite strong. The results highlight the effectiveness of models using combined multimodal data under challenging conditions, which agrees with the results from \cite{liu2020simultaneously}.

\subsection{Comments on Selective Pose Prior Transfer} 
\label{sec:prior}

HW-HuP learning is guided by a source prior that fades as training on the target data progresses. Our ablation study shows the effectiveness of this feature when the SLP dataset is the target. One might wonder how such a mechanism can be effective---why not just learn directly from the target domain, which has a different pose distribution? In our experience, the source prior contribution mitigates the effect of \textit{incompleteness} of the target domain due to limited observations and data. For example, the target domain might contain just enough information about major joint locations to regulate the placement of major body parts, but more subtle details such as axis rotations and the placement of small body parts are relatively unconstrained. 

The visual examples in \figref{sotaPre} show that models pre-trained on widely available human pose datasets tend to produce estimates of human poses that appear to be standing or squatting, even though the input images are of humans lying in bed. The main improvements from the HW-HuP estimate come from adjustment to major limbs, while rotation and placement of the head and hands are largely inherited from the source priors. 

We can think of the source prior estimation and the true target pose as neighbours in the human pose manifold, with strong similarities in the small limbs. A pose guess from a source prior to regress to the target pose is always better than a guess from nowhere and we hope the evolution follows such a manifold. With a strong prior at beginning, the regressor $F$ learns under the source prior guidance to learn the overall picture of full joint pose (rotation). As the source prior fades away, $F$ will focus more on the observations from target domain to have its context-specific prior over the major joints yet keep the learned prior over the minor ones which is otherwise unconstrained under the target-only supervisions. In this way, the source pose prior is actually selectively transferred to the target regressor $F$.

Another benefit of the selective pose prior transfer, compared to transfer learning by continuing training of a pre-trained model, is that it is still helpful when the source and target domains use different input modalities, and so no pre-trained model in the target modality exists. Since the learned pose prior is modality independent, it can be used to guide learning even under a new modality.

\subsection{Error Distribution from depth-based 3D proxy}
\figDpErr
\figref{dpErr} shows the error distributions resulting from using depth-based 3D proxy coordinates as estimates of the real proxy coordinates, in the training split of Human3.6M. We can see that biases tend to be larger at more distal joints, and smaller at proximal joints. This is in accordance to the our assessment in the main text, that double- or multi-body occlusions are more likely at distal joints, leading to higher Type 2 errors. 

\subsection{Comprehensive Comparison with SOTA on Human3.6M}

\tabref{humanCall} is a more comprehensive version of \tabref{humanD}, comparing HW-HuP with SOTA predictions on the Human3.6M.

\tblHumanCall
\subsection{More Qualitative Results} 
\figAllCsup

We exhibit further 3D prediction examples from the SLP dataset in \figref{allCsup} and from the Human3.6m dataset in \figref{hmsup}. In \figref{allCsup}, we include predicted results based on different input modalities (depth, LWIR, PM, or multi-modal) for subjects under different cover conditions (no cover, a thin layer sheet, or a thick blanket). Some failure cases are included such as in row 1, column 1, where LWIR fails, and also in row 3, column 1, where the left hand is not in a rest pose. 
For the visualization of Human3.6m frames in \figref{hmsup}, we observe that although the performance of our HW-HuP model is comparable to the SOTA 3D pose estimation models, there are still some failure cases. For instance, the predicted upper limbs are more difficult to align to the point cloud than the lower limbs (e.g. as in the second pose of subject 9 and the fourth pose of subject 11).

\fighmsup

}

\begin{document}
%\sloppy
\title{Heuristic Weakly Supervised 3D Human Pose Estimation}
%\subtitle{Do you have a subtitle?\\ If so, write it here}

%\titlerunning{Short form of title}        % if too long for running head

\author{Shuangjun Liu    \and Michael Wan \and
        Sarah Ostadabbas %etc.
}

%\authorrunning{Short form of author list} % if too long for running head

\institute{S. Liu,  M. Wan and S. Ostadabbas \at
              Augmented Cognition Lab, Electrical and Computer Engineering Department, Northeastern University.
              %\email{shuliu@ece.neu.edu ,sehgal.n@husky.neu.edu,ostadabbas@ece.neu.edu}           %  \\
}

\date{Received: date / Accepted: date}
% The correct dates will be entered by the editor

\maketitle

\begin{abstract}
Monocular 3D human pose estimation from RGB images has attracted significant attention in recent years. However, recent models depend on supervised training with 3D pose ground truth data or known pose priors for their target domains. 3D pose data is typically collected with motion capture devices, severely limiting their applicability. In this paper, we present a heuristic weakly supervised 3D human pose (HW-HuP) solution to estimate 3D poses in when no ground truth 3D pose data is available. HW-HuP learns partial pose priors from 3D human pose datasets and uses easy-to-access observations from the target domain to estimate 3D human pose and shape in an optimization and regression cycle. We employ depth data for weak supervision during training, but not inference. We show that HW-HuP meaningfully improves upon state-of-the-art models in two practical settings where 3D pose data can hardly be obtained: human poses in bed, and infant poses in the wild. Furthermore, we show that HW-HuP retains comparable performance to cutting-edge models on public benchmarks, even when such models train on 3D pose data\footnote{Our model code and data are publicly available at \url{https://github.com/ostadabbas/hw-hup}.}.

\end{abstract}

%-------------------------------------------------------------------------
\section{Introduction}
%The small data problem becomes critical as machine learning (ML) techniques grow more data hungry. In the past, the small data problem was an outlier and not of particular interest to the general ML community, due to the availability of some large-scale datasets \citep{russakovsky2015imagenet}. However, in the last few years, 

Applied machine learning (ML) researchers have repeatedly come up against the problem of data scarcity in practical settings. Data-efficient methods, which exploit structural knowledge in each domain to constrain models to being simple enough to train with the available data \citep{dai2007boosting}, have emerged in response. However, data-efficient ML often requires at least \textit{some} sample data from the target domain, or, in the case of zero-shot learning \citep{romera2015embarrassingly}, some shared attributes in the source and target domains. But what can be done if context constraints entail that not even a single annotated sample is available for a target application?
%Though Unsupervised Domain Adaptation (UDA) is not a new topic \citep{pan2009survey}, the discussions are mainly about the image and feature level alignment instead of 3D human pose prior. 

\figSota
% Let's look at the problem of 3D human pose estimation in applications in the a real application especially under a novel context, where no ground truth 3D pose data are accessible, neither any known priors. 
% This paper 
Our paper addresses this challenge for the task of 3D human pose estimation in the wild, where no ground truth 3D pose data are accessible and known priors do not exist. We present a transfer learning approach based on a heuristic weakly supervised framework with depth data as surrogate 3D information. Our work focuses on applications where techniques based on the motion capture (MoCap) systems---which involve specialized markers, multiple cameras, and controlled environments---are not feasible, even for gathering training data. These contexts might include hospital patient behavior monitoring, as studied by our lab \citep{liu2020simultaneously} and others \citep{clever2020bodies}, or infant in the crib activity monitoring \citep{hesse2017body}. \figref{sota} demonstrates the qualitative limitations of state-of-the-art (SOTA) 3D pose estimation models for in-bed patient monitoring---an application critically in need of automation, as highlighted by the recent global pandemic.

% Though numerous works have been proposed in recent years via depth-based sensing, their focus are mainly on reconstructon of depth, segmentation or surface normal via monocular/stereo vision signal \citep{wang2019pseudo, xu2018pad, ma2019accurate, vandenhende2020mti, zhang2019pattern}. We should be aware that knowing depth or surface normal does not equals to 3D human pose. These efforts do not specifically address the ground truth missing problem of 3D human pose estimation. 

%It is not trivial as the ultimate goal of all the community effort is making these models into practical tools to change our daily lives instead of hitting a benchmark.   Though impressive studies are emerging over 3D Human Pose estimation problems, yet mainly the main focus competing each other over existing benchmarks,  how to introduce or repurpose them into a practical environment is seldomly discussed. 

In place of 3D data from MoCap, we assume that training data is captured with an inexpensive, off-the-shelf RGBD camera, which adds depth information to 2D RGB data \citep{zimmermann20183d}. 
% Note that the additional depth data is imprecise, even after processing with popular pose estimation software (e.g. the Microsoft Kinect SDK). 
Note, that RGBD data does not directly provide 3D joint locations, as the depth cloud generally lies on the surface of the body, and this discrepancy is exaggerated for limbs occluded by other body parts. Existing approaches offer plausible solutions for human mesh tracking and recovery from RGBD data, but typically have requirements which limit their general applicability: \citep{hassan2019resolving} needs a static scan of the environment; \citep{hesse2019learning} assumes that the subject is lying on a flat, decluttered surface, and requires sequential RGBD data for training; \citep{bogo2015detailed} requires a static background where the subject point cloud can be cleanly segmented, and also sequential data; and \citep{zimmermann20183d} requires depth data during inference. By contrast, our method is trained on single RGBD images, makes inferences on single RGB images, and is free from environmental requirements. Our heuristic weakly supervised 3D human pose estimation model (HW-HuP) is trained using both the learned pose and shape priors from public 3D pose datasets, as well as 2D pose and depth observations from the target domain. By partially learning the priors from the source domain and the noisy observations from the target domain, our approach iteratively converges to a reliable 3D pose estimation. The final product is a predictive model which can estimate 3D human pose and shape directly from a single 2D image taken in the wild (camera calibration parameters are neither needed nor assumed). 

Our results show that HW-HuP performs robustly in real-world scenarios where 3D pose data is not available, such as with individuals in bed, or infants moving freely in cribs or playrooms. We also show that HW-HuP retains strong performance under different input modalities, such as pressure map, depth, or infrared signals. Finally, HW-HuP exhibits performance comparable to SOTA methods on public benchmarks in the lab and in the wild, even when those methods take advantage of full 3D ground truth data in training and HW-HuP does not.

\section{Related Work}
% Here, we give a brief overview of the existing work on 3D human pose estimation where they are supervised either by MoCap data or depth data. \com{we need a short overview paragraph here. what is the relationship of weakly supervised and unsupervised ones to other subsection?}
% Since 3D human pose estimation under a novel context is our major focus, here we give a brief overview of the existing work on 3D human pose estimation where they are supervised either by MoCap data or RGBD/depth data.

% The 2D human pose estimation problem has been extensively studied in the past couple of decades based on 2D images, however when it comes to 3D pose estimation, addition 3D information is often necessary for model supervision, which usually  comes from MoCap or depth sensing devices. 

%we give a brief overview of the existing work we will first introduce relevant works for 3D human pose estimation. Secondly, as depth is employed as our supervision signal and there exist depth based approaches which are able to track human motion and even recover detailed human mesh at runtime. So we will specifically talked about the depth or RGB-D based approaches and their limitation in our applications.   

\textbf{RGB-based 3D Human Pose Estimation:}
%\subsection{3D Human Pose Estimation:}
While we focus our attention here on 3D human pose estimation from RGBD images, there has naturally been extensive work on such estimation from RGB images. These run the gamut from end-to-end methods \citep{mehta2017monocular,park20163d,pavlakos2017coarse,rogez2017lcr}, to lifting 2D estimations to 3D \citep{zhou20153d,chen20173d,tung2017adversarial,rhodin2018unsupervised}. Our model draws on work from \citep{bogo2016keep}, which directly regresses the pose and shape parameters from a learned template.

\textbf{RGBD-based 3D Human Pose Estimation:} Some techniques successfully use RGBD data to predict human poses, including detailed body meshes \citep{de2007marker,bogo2015detailed,li2019towards}. The popular Microsoft Kinect RGBD camera even includes a 3D pose estimator in its software development kit (SDK), so it is tempting to conclude that RGBD data is sufficient for 3D human pose estimation in general. However, in practice, these methods have significant constraints, restricting their application to controlled environments. For example, for full body mesh reconstruction, continuous depth data is often required; initialization poses, cleanly segmented point clouds, are sometimes needed in \citep{helten2013personalization,ye2014real,lan2012double,yu2017bodyfusion,hesse2019learning}; and static scans of the empty environment are requisite in \citep{hassan2019resolving}.
% The SMIL model from \citep{hesse2019learning} tackles the problem of transfer to the new domain of infant bodies, but their training method is highly constrained, requiring cleanly segmented point clouds, continuous (video) image frames for training, and a specialized flat surface environment. 
% In particular, the method in \citep{hesse2019learning} 
% their data term requires clearly segmented point cloud to fit the mesh; The temporal smoothness term requires the continuous frames; The table term employs the specific constraint from the truth of the baby is on a flat table surface... these models aim at high quality pose and shape tracking without possible interference from the natural environment. By contrast, our work focus on the fine tuning of existing 3D pose models without aforementioned constraints and can be easily introduced for novel application even from a wild setting.
Single frame depth estimation approaches do exist, but they fail on novel applications \citep{liu2020simultaneously} and still require ground truth 3D pose or segmentation annotations \citep{shotton2011real,haque2016towards,xiong2019a2j}.

\section{Method}
\label{sec:PF}
This paper focuses on 3D human pose estimation when for training only 
%As we define our problem as is fine-tuning of 3D human pose model for specific application when no ground truth 3D pose is available, to focus on the missing link, 
RGBD data is accessible and during inference only RGB data is available. In order to separate our 3D pose estimation task from the 2D problem, we assume that the 2D poses are given---either from an effective 2D pose estimator, or refined with human input from a baseline 2D pose estimator using the AI human co-labeling toolbox \citep{huang2019ah} developed by our lab.

In sum, the following \textit{data availability} is assumed: (1) 2D image data $I$ (RGB, unless otherwise specified); (2) depth data $D$; (3) a 2D pose ground truth annotation $x_\text{gt}$; and (4) a 3D pose estimation model $F$ pre-trained on large-scale publicly-available 3D human pose datasets. Since we are targeting applications in natural settings, we also assume the following \textit{data constraints} on the depth $D$: (1) $D$ is available during training but not inference; (2) $D$ may not be continuous across frames; and (3) $D$ might be so noisy that the subject cannot be easily segmented from the background.
% especially since the existing RGBD cameras have limited range and anything outside that range is not sensed reliably. Also, if the target body is too close to the background, the depth detection performance drops dramatically, which is the case in MS Kinect SDK. 

% Due to the data constraints in the target domain, we can take advantage of some priors learned from the existing 3D human pose datasets.

\figDepthProxy
% \textbf{Uncertainty in Depth Observation:} We can combine the 2D joint information $x$ and the depth $D$ to obtain \textit{depth-based proxy 3D coordinates} $X_\text{dp}$, but these can differ from true joint locations. 
By combining 2D joint $x$ and the depth $D$, we have \textit{depth-based proxy 3D coordinates} $X_\text{dp}$, which is actually different from the true joint location. 
Setting aside irreducible error stemming from the noisiness of $D$, there are two fundamental and nonuniform sources of bias for these depth-based proxy 3D coordinates, which depend on pose and shape. %When a human subject stays in a canonical pose, this bias is usually negligible since it is limited by the limb thickness.
% If there were equal bias across all body joints, the true 3D pose could still  be achieved by making everything root centered. However, these biases are  not equal or constant and they are both pose- and shape-dependant. 
% A large limb does not necessarily hold the same bias as a thin limb due to their shape differences. The bias may also vary from different view points caused by pose differences. 
% For a better illustration, we assume the limbs hold oval cross section and illustrate the bias varying effect in \figref{DepthProxy}. 
%
First, proxy 3D coordinates will sit on the surface of a body, whereas the true 3D skeletal joint coordinates reside inside the body. Critically, this offset may not be uniform across joints due to shape and pose differences, as illustrated for the idealized human limb in \figref{DepthProxy}(a). We refer to the resulting error as a \textbf{Type 1 error}. Another source of bias stems from body parts overlapping in such a way that one joint is occluded from the camera by another joint or body part, as with crossed arms or legs, and as illustrated ideally in \figref{DepthProxy}(b) and more concretely in \figref{DepthProxy}(c). This error is usually large across limbs, and is referred to as  \textbf{Type 2 error}.
%The type 2 error is usually correlated to the occlusion such as the end body limbs such as wrists and ankles. Therefore, the  3D joint location estimated based on the depth data is only a proxy from the true 3D pose. 
Type 1 and Type 2 error distributions for each joint are provided in the \textit{Supplementary Materials}.

% We use $d_i(\theta, \beta)$ for the bias from true joint $i$ position to its proxy location. In most cases $d_i(\theta, \beta) \neq d_j(\theta, \beta)$ for different joint i and j. 
%We call the estimated 3D joint location  in this approach the 3D joint depth-based proxy. 
% However, exact modeling these bias are not trivial. 

% \subsection{Using Depth Data}
% \com{I really do not get what is this paragraph:} Another observation we have is the depth $D$, however it is not straight forward to introduce for supervision as there are no exact mapping from recovered point cloud to the template vertices. 
% Under a natural setting, estimation of such mapping is challenging due to the depth data could be discontinuous and noisy and the template is deformable and articulated. 

%\section{Introducing HW-HuP} 
% The data distribution gap between the public MoCap data and a practical pose estimation application often stems from the differences in types of poses that the subjects take in different domains. As a result, the pre-trained SOTA  models do not show satisfactory performance on a novel application unless being fine-tuned. Nonetheless, having no ground truth 3D pose data prevents a straightforward fine-tuning. 

\subsection{HW-HuP Problem Formulation} 
Our objective is to obtain a 3D pose regression model $F$ which takes in a single 2D image $I$ and estimates 3D human pose parameters $\Theta = [\theta, \beta]$ and camera parameters $C$, via $F(I)=[\Theta, C]$. Following \citep{kolotouros2019learning}, we use the skinned multi-person linear model (SMPL) \citep{loper2015smpl}, with pose  $\theta \in \mathbb{R}^{3K+3}$ representing the relative rotation angles of $K=23$ body parts with respect to their parents in the kinematic tree, plus the root global rotation, and shape $\beta \in \mathbb{R}^{P}$ representing the first $P=10$ principle components analysis (PCA) coefficients of the human template space. The SMPL model is a differential function that outputs a triangulated mesh $M(\Theta) \in \mathbb{R}^{3 \times N}$ with $N=6980$ vertices. The camera model $C=[T, s]$ includes 
% camera global rotation $R\in \mathbb{R}^{3 \times 3}$, 
a translation vector $T \in \mathbb{R}^2$ and a scale $s \in \mathbb{R}$. For a given pose $\Theta$ (sometimes implicitly including $C$), each 3D keypoint $X(\Theta)\in\mathbb{R}^3$, is a fixed linear combination of the vertices from the mesh $M(\Theta)$ \citep{kolotouros2019learning}. Its 2D projection $x$ is given by $x = s \Phi (RX(\Theta)) +T$,
where $\Phi$ is an orthographic projection and $R$ is the global rotation matrix $R\in \mathbb{R}^{3 \times 3}$, both derived from $C$. 
% For each symbol, we add hat to represent the estimated version. 

\subsection{HW-HuP Model Description}
HW-HuP extends SPIN \citep{kolotouros2019convolutional},
% which estimates 3D human poses from 2D images via a collaborative regression--optimization framework,
by incorporating depth data into the supervision in two stages in a coarse-to-fine manner.
% respectively targeting the two types of depth errors discussed earlier.
HW-HuP assimilates source priors differently, and allows for input modalities beyond RGB. Recapping the SPIN work, we have the human pose and shape regression model $F$ with output of SMPL parameters $F(I) = \Theta_\text{reg}$ and corresponding 2D joints $x_\text{reg}$ might be trained with supervision from ground truth 2D joints $x_\text{gt}$ via the loss function
\begin{equation}
    L_\text{2D}(\Theta_\text{reg}) := \sum_{\text{joints } j}\left\Vert x_\text{reg}(j) - x_\text{gt}(j)\right\Vert_2^2.
\end{equation}
The key idea behind SPIN is to instead use the ground truth 2D joint data $x_\text{gt}$ to fit the model $M(\Theta_\text{reg})$ to the image $I$ via an iterative optimization procedure called SMPLify \citep{bogo2016keep}, and then using the resulting optimized model parameter $\Theta_\text{opt}$ to supervise $F$ via the loss function
\begin{equation}
    L_\text{SPIN}(\Theta_\text{reg}) := \left\Vert \Theta_\text{reg} - \Theta_\text{opt}\right\Vert_2^2.
\end{equation}
The loop $I\mapsto\Theta_\text{reg}\mapsto\Theta_\text{opt}$ represents a single SPIN forward step, and with mechanisms in place to discard bad optimization results, both the regression $F$ and the SMPLify optimization improve ``collaboratively'' over time.
%\footnote{In deference to the structure of the SPIN model, we reserve the term ``regression'' for the pose estimator $F$ and ``optimization'' for the SMPLify fitting algorithm.}

HW-HuP modifies SPIN by tweaking the SMPLify optimization procedure, and more importantly, by incorporating depth data into the supervision of the regression $F$ in one of two ways depending on the training epoch, resulting in an overall loss function of
\begin{align}
    &L_\text{reg}(\Theta_\text{reg}) :=  L_\text{2D}(\Theta_\text{reg}) + L_\text{SPIN}(\Theta_\text{reg})\\ &\phantom{L_\text{reg}(\Theta_\text{reg}) := } + \begin{cases} 
    L_\text{3D}(\Theta_\text{reg}) & \text{ in Stage I} \\  L_\text{depth}(\Theta_\text{reg}) & \text{ in Stage II} \nonumber\end{cases}
\end{align}
(with $L_\text{2D}$ included following the SPIN implementation). We now turn to detailed descriptions of each modification that HW-HuP makes to SPIN with an overview shown in \figref{HWS}.
% and define the depth losses $L_\text{3D}$ (coarse) and $L_\text{depth}$ (fine). See \figref{HWS} for an overview.

\figHWS

\textbf{Selective Pose Prior Transfer (Optimization):} The iterative optimization used in SPIN is a modified version of the SMPLify algorithm \citep{bogo2016keep}. It takes the regressed pose prediction $\Theta_\text{reg}$ (here, including $C_\text{reg}$) as initial input and outputs an optimized $\Theta_\text{opt}$ after a number of internal steps, all within a single forward pass of each SPIN training step. The modified SMPLify algorithm first optimizes the camera parameter $C_\text{reg}\mapsto C_\text{opt}$, and then optimizes $\Theta_\text{reg}\mapsto\Theta_\text{opt}$ via a multi-step iterative process guided by the objective loss function
\begin{align}
    &L_{\text{opt}}(\Theta_\text{opt}) := L_{\text{2D}}(\Theta_\text{opt}) + \lambda_\theta L_\theta(\theta_\text{opt}) \\ 
    &\phantom{L_{\text{opt}}(\Theta_\text{opt}) :=} + \lambda_\beta L_\beta (\beta_\text{opt}) +  \lambda_\alpha L_\alpha (\theta_\text{opt}),\nonumber
\end{align}
where $L_{\text{2D}}$ is the 2D joint loss, $L_\theta$ is GMM pose prior learned from the source domain, $L_\beta$ is a quadratic penalty on the shape coefficients, $L_\alpha$ is an unnatural joint rotation penalty of elbows and knees, and $\lambda_\theta$, $\lambda_\beta$, and $\lambda_\alpha$ are scalar constants.
%\footnote{Note that many steps of this internal SMPLify optimization occur within a single training step for $F$, so $L_\text{opt}$ and $L_\text{reg}$ are quite distinct and cannot be combined, although they share the $L_\text{2D}$ term.}
HW-HuP 
% uses the same optimization procedure,but 
modifies the coefficient $\lambda_\theta = \lambda_{\theta,0} f^e$ to introduce an exponential decay of strength $f$ over epoch $e$ from the initial $\lambda_{\theta,0}$ weight for the GMM prior. As the source prior fades over time, the major body parts tend to become increasingly guided by incoming target data, but the influence the source prior is retained for small body parts.
% (More discussion is provided in \textit{Supplementary Materials} \secref{prior}). 

%- from supp 
% comments on pose prior

\textbf{Coarse 3D Pose-Based Supervision (Stage I):} In Stage I, we want to supervise the regression $F$ by penalizing differences between its 3D joint predictions $X_\text{reg}$ and the true 3D joint locations, but these are unknown. What we do have are the depth-based proxy 3D coordinates, $X_\text{dp}$, but these can be biased away from the true 3D joint locations, with significant variation as illustrated in \figref{DepthProxy}. 
To remedy the most egregious Type 2 errors stemming from occlusion, we design and train a visible joint detection model $V$ and use it to construct our weakly supervised 3D pose loss:
\begin{equation}
    L_\text{3D}(\Theta_\text{reg}) = \sum_{\text{joints } j} V(j)\left\Vert X_\text{reg}(j) - X_{\text{dp}}(j)\right\Vert_2^2,
\end{equation}
where $V(j)$ is the visibility for joint $j$. The idea is that in cases of heavy occlusion of $j$, a low visibility prevents $X_\text{dp}(j)$ from sending the potentially wrong signal in the backward pass. The regressor $F$ instead learns the 3D location of that occluded joint by inferring from nearby poses where that joint is more visible. Our visible joint detection model $V$ is obtained by fine-tuning a SOTA pose estimation model and adding a visible joint detection branch called VisNet. Details are provided in \textit{Supplementary Material}. In cases where visibility annotations are provided for 2D joints
% \citep{andriluka20142d,lin2014microsoft,liu2020simultaneously}, 
we use them directly for $V$.
% however the information is not usually used since most  2D pose estimation models attempt to infer all joints in the image no matter they are visible or not. Nonetheless, 
%In $X_{\text{dp}}$ supervision, it is critical to avoid the significant bias in the observations.  
% hope we can have this one 
% elaborated in detail in \secref{details}, on top of its backbone network.
% Details are provided in our \textit{Supplementary Materials} \secref{visnet}.
% VisNet head is designed to be a two $1\times1$ convolution layer followed by three fully connected layers to solve a  multiple binary classification problem of multi-joint visibility. Each layer is followed by a batch normalization and a ReLU. VisNet .

% \textbf{Observation Interpretation at Run-Time}
\textbf{Fine Depth-Based Supervision (Stage II):}
In Stage I, the 2D joint data $x$ and source prior supervision ensure that our model yields a plausible frontal view with aligned 2D poses, and the 3D proxy joint $X_\text{dp}$ supervision ensures that the estimated joints will be close to their true locations in $3D$ space, with most Type II errors reduced.
However, small bias introduced by % the proxy supervision means that 3D joint estimates $X_\text{reg}$ will be located on the surface of the body (near $X_\text{dp}$), rather than inside of it, so 
Type I errors still remain. In Stage II, we take advantage of successful 2D and 3D alignments in Stage I to enable fine-grained depth supervision with the full depth data $D$ to further refine the estimation $F$ and reduce Type I errors. To do this, we employ a differential neural renderer (NR) \citep{kato2018renderer}, which takes in the estimated 3D body mesh $M(\Theta_{\text{reg}})$ and returns the corresponding depth map $D_\text{NR}$ to align with the observed depth $D$. 
% which again, at the beginning of Stage II, tends to exhibit Type I errors compared to $D$. However,
After stage I alignment, the discrepancies only exist over the silhouette of the body, thus, we also make use of the estimated silhouette, or 2D mask $\mathcal{M}=\mathcal{M}_\text{NR}$, which is provided by the neural renderer. If a prior mask $\mathcal{M}_\text{prior}$ is available---such as an annotated ground truth mask or a weak estimate like a bounding cubic indicating the foreground area---we work with its intersection with the neural renderer prediction: $\mathcal{M}=\mathcal{M}_\text{NR}\cap\mathcal{M}_\text{prior}$. The fine depth loss for Stage II is thus
\begin{equation}
    \label{eqn:depth-error}
    L_{\text{depth}}(\Theta_\text{reg}) = \left\Vert D - b_0 - D_\text{NR} \right\Vert^2_{2, \mathcal{M}},
\end{equation}
where $\left\Vert\cdot\right\Vert_{2, \mathcal{M}}$ is the $L^2$ norm taken over pixels in $\mathcal{M}$ only, and $b_0$ is a correction factor for the bias introduced by using imaginary focal length, 
which minimizes the average distance from the observed depth $D$ to the rendered depth $D_\text{NR}$:
\begin{equation}
    b_0 = \argmin_{b} \left\Vert D -b- D_\text{NR}\right\Vert_{2, \mathcal{M}}.
\end{equation}
Since our $D$ is smoothed and filtered, we use the $L^2$ norm in $L_\text{depth}$, but in cases of noisy $D$ or with many outliers,  we recommend using a robust penalty loss such as German-McClure \citep{geman1987statistical} instead. 

\section{Experimental Evaluation} 
\subsection{Evaluation Datasets}
% this para may cut 
We evaluate the performance of the HW-HuP framework with the following datasets.
%by using no ground truth 3D pose data during training. 
% when no ground truth 3D pose data is accessible even during  training. 
%The first dataset is from a practical application with no 3D pose data, but it includes RGB, depth, and other imaging modalities for each pose. The second dataset has 3D ground truth information as well as depth data, which allows us to quantitatively compare our  HW-HuP framework with the other SOTA 3D pose estimation models.  
The first three are used to quantify real-world performance, in the absence of 3D pose annotations or priors.

\textbf{SLP Dataset:}  
% We first focus on an emerging healthcare application for in-bed pose monitoring \citep{liu2020simultaneously}. We use 
The simultaneously-collected multimodal lying pose (SLP) dataset \citep{liu2020simultaneously} was collected with IRB approval by our team under Northeastern University approval IRB \#17-06-04, from seven participants in a hospital room and 102 participants in a living room. The four modalities of RGB, long wavelength infrared (LWIR), depth, and pressure map were simultaneously collected under three cover conditions: no cover (``Nocover''), a thin layer sheet (``Cover1''), and a thick blanket (``Cover2''). SLP does not have 3D pose annotations and exhibits a pose prior distinct from those found in popular pose datasets. Correspondingly, pre-trained SOTA models perform weakly on SLP \citep{liu2020simultaneously} and there are limited works addressing the 3D human pose problem presented in SLP dataset \citep{clever2020bodies}, making it a useful real-world testing ground for our framework.
% We begin with a commonly-used RGB modality in order to compare our  3D pose estimation performance with other SOTA approaches. Then, we extend this evaluation to more challenging cases, where the human body is heavily occluded in the RGB images. We introduce  other input modalities beyond RGB such as pressure map, LWIR, or depth to show HW-HuP general effectiveness in 3D pose estimation.
Since there is no ground truth 3D pose data, we measure performance using the \textit{aligned depth error}, defined as the mean of the distances (in mm) between the predicted $D_\text{reg}$ and the provided ground truth $D=D_\text{gt}$, taken over the intersection of the predicted and ground truth silhouette masks, after camera alignment.

\textbf{SyRIP} \citep{huang2021infantpose}: The synthetic and real infant pose (SyRIP) dataset, also developed by our team, combines the real (web-sourced) and synthetic infant images. We use the first 1600 images for training and the remaining 100 for testing.

\textbf{MIMM}: This privately-owned modeling infant motor movement (MIMM) dataset was collected by a medical startup company with collaborative institutional approval (as part of an NIH funded study), and contains video recordings of interactive motor assessment sessions from 68 infants under age one with their caregivers and clinicians. Depth data from an MS Kinect camera is provided. 

The next two datasets are used to quantify the error in HW-HuP introduced by not using 3D pose ground truth data.

\textbf{Human3.6M} \citep{ionescu2013human3}:
A large-scale indoor daily activity MoCap benchmark for 3D human pose estimation. We employ Protocol\#2 of the benchmark, where subjects S1, S5, S6, S7, S8 are used for training, and S9, S11 for testing.
% In this test, we employ the mean per joint position error (MPJPE) and also the Procrustes analysis (PA) MAJPE \citep{sun2018integral,ionescu2013human3,gower1975generalized} as the evaluation metrics. 

\textbf{3DPW} \citep{vonMarcard2018}: This dataset features mostly outdoor environments with poses captured via the Inertial Measurement Unit (IMU) sensors, and thus serves as a good benchmark for performance in the wild. We use it for evaluation only.

\subsection{Implementation Details}  % may be put in supplementary. 
\label{sec:details}
\textbf{Data Preparation:} Depth data is preprocessed with denoising \citep{nguyen2012modeling} and hole-filing \citep{liu2016computationally} algorithms. The joint depth proxy $X_{\text{dp}}$ is extracted from the 2D joint location $x$  from the denoised depth image and reprojected into 3D in the camera space. 
All images are normalized cropped to center bodies. Random flipping, rotation, scaling, and random channel-wise noise are used for data augmentation. 

\textbf{Network Training:} For the 3D regression $F$, we take the same design and initialization as in \citep{kolotouros2019learning}. We employ the differential neural renderer NR in  \citep{kato2018renderer}. We follow the SMPLify and SPIN optimization loops used in \citep{kolotouros2019learning,bogo2016keep} and limit the max iterations to 50. All models are trained on an 
NVIDIA Tesla K40m with batch size 64.  We employ Adam \citep{kingma2014adam} with learning rate 5$\times$ 10$^{-5}$. For the SLP dataset, the model is trained in 30 total epochs, with Stage II starting at the tenth epoch. 
% The training takes about 20 hours for uncovered RGB and 40 hours for all cover conditions. 
For Human3.6M, due to the high frame quantity and the near-repetition of poses in neighboring frames, we employed a downsampling rate of 50.
% and limit the iteration for each epoch to 150. 
Training takes four total epochs, with Stage II starting at the second epoch. 
For our infant model, we combine the SyRIP and MIMM datasets and train for a total of 640 epochs with Stage II starting at epoch 420.
% VisNet is trained on COCO \citep{lin2014microsoft} with Adam optimizer and learning rate 0.001.

%Implementation details and performance are provided in \textit{Supplementary Materials} \secref{visnet}.
% visnet intro 

% Qualitative result of VisNet on COCO and Human3.6M \citep{ionescu2013human3} are shown in \figref{visnetGrid}. 
% We notice that it is more likely to find the occluded joint when the occluded object has distinct appearance from the blockers. 
% We also notice some failures cases as the following: (1) When the joints are close to the image boundary or other objects. We assume it may depend on the appearance differences as boundary to other objects or image edge all produce sharp changes such as \figref{visnetGrid} row(r) 2, column(c) 6. 
% (2) Local view of the limbs, see (r1, c6). In the  lower body only cases, it is hard to learn a good inference and most parts may be deemed as invisible. 
% (3) If one of the joints is invisible, the neighboring joints on the same limb are sometimes deemed as invisible, see r2 c7.  

\figRGBuc 
\subsection{Results and Ablation Study}
\label{sec:abalation}
We chose the SLP in-bed pose dataset to illustrate the effectiveness of HW-HuP in a domain with distinct and previously unknown pose priors. We start by considering images in the RGB modality where the subject is not covered by a bed sheet, to most closely match the testing conditions employed by other SOTA models \citep{kolotouros2019learning,kanazawa2018end}.
For a fair comparison, we compare models which share the same input modality. This is also helpful to compare their general effect on different dataset later. 
% Although exact 3D pose estimation performance cannot be calculated since ground truth 3D  data is missing, we provide the aligned depth error (AD-e) as a reference. 
We report the aligned depth error and also display some predictions in figures, across various configurations, to highlight qualitative differences in performance.
The SOTA models we compare with are:

\begin{itemize}
    \item \textbf{SPIN*}: Pretrained model from \citep{kolotouros2019learning}.
    \item \textbf{SPIN}: SPIN* fine tuned on SLP 2D pose training data.
    \item \textbf{HMR}: Pretrained model from \citep{kanazawa2018end}, fine tuned on SLP 2D pose training data.
\end{itemize}

% the pre-trained model SPIN* \citep{kolotouros2019learning}, its fine-tuned version SPIN on SLP, and the fine-tuned  human mesh recovery (HMR) model \citep{kanazawa2018end} on SLP. 

We define configurations following simple plausible assumptions for our ablation study:

\begin{itemize}
    % \item \textbf{3D-dp (SMPLify + partial Stage I)}: We supervise HW-HuP with just $x_\text{gt}$ and $X_{\text{dp}}$, skipping Stage II entirely. This would be effective in domains where 2D poses and depth data together yield depth-based proxy 3D joint coordinates $X_{\text{dp}}$ which match the ground truth 3D coordinates.
    % \item  \textbf{3D-dp-vis (SMPLify + Stage I)}: We supervise HW-HuP with just $x_\text{gt}$, $X_{\text{dp}}$, and visibility data $V$, skipping Stage II entirely. This would be effective in domains where 2D poses and depth data together yield depth-based proxy 3D joint coordinates $X_{\text{dp}}$ which match the ground truth 3D coordinates, except in cases of occluded joints.
    % \item \textbf{3D-dp-vis-D (full HW-HuP)}: The full HW-HuP full model.
    % \item \textbf{noPrior}: We set the source prior constraint to zero. This would be effective in domains where the source prior is misleading.
    % \item \textbf{2D-D (Stage II only)}: We remove the SMPLify optimization loop entirely, and instead supervise $F$ with only $x_\text{gt}$ and $D$. This effectively fine-tunes $F$ directly with the available data.
    \item \textbf{3D-dp (SMPLify + partial Stage I)}: \textit{Assumption: 3D depth proxy is equivalent to the real 3D pose.} Supervise with just $x_\text{gt}$ and $X_{\text{dp}}$, skipping Stage II. 
    \item \textbf{3D-dp-vis (SMPLify + Stage I)}: \textit{Test whether visibility helps by reducing the Type II errors.} Supervise with $x_\text{gt}$, $X_{\text{dp}}$, and visibility data $V$, skipping Stage II. 
    \item \textbf{3D-dp-vis-D (full HW-HuP)}: The full HW-HuP model.
    \item \textbf{noPrior}: \textit{Assumption: The source prior is not needed at all to train a model for target domain.} Set source prior constraint to zero. 
    \item \textbf{2D-D (Stage II only)}: \textit{Assumption: Fine-tuning directly with the available data is just as effective.} Remove SMPLify optimization loop; instead supervise $F$ with only $x_\text{gt}$ and $D$. 
    % This effectively fine-tunes $F$ directly with the available data.
\end{itemize}

% \figVisnetGrid

\tblAbla
All models are trained with SLP ``Nocover'' RGB data, except the pretrained SPIN* model from \citep{kolotouros2019learning}. The resulting aligned depth errors (in mm) are reported in \tabref{abla}. Note that this metric can only partially represent the estimation quality as it cannot accurately quantify positions of occluded limbs. The results show that fine-tuning on the SLP data significantly improves SPIN performance compared to the purely pretrained SPIN*, but even then, SPIN and HMR perform notably worse than all HW-HuP models. Our ablation results show that depth proxy supervision (dp) by itself brings HW-HuP ahead of SPIN and HMR performance. Other model features such as the visibility supervision (V), full depth supervision (D), and source priors all further improve the model, with the best performance obtained by the full 3D-dp-vis-D. The good performance of the 2D-D configuration shows that straightforward supervision already leads to acceptable estimation, but in modalities other than RGB where no strong pretrained model is available, 2D-D performance is more limited.
% However, this is only the case when we have RGB input with no occlusion. The 3D estimation performance degrades in 2D-D case,  when we extend HW-HuP beyond RGB, where no pre-trained model exists. 

% \figRGBuc 
Qualitative results for our experiment are presented in \figref{RGBuc}, which focuses on typical settings and representative issues. We highlight the tendency of SPIN \citep{kolotouros2019learning} to adhere to the source prior for daily activities, as exemplified by the hands sticking out in supine and straddled legs in the side lying positions. 3D-dp is generally satisfactory for supine poses, which make sense since these poses typically exhibit less limb occlusion, eliminating the source of the largest depth proxy biases; but local defects like the slightly raised head and the legs bent into the bed remain. By contrast, all of these errors are handled by the full 3D-dp-vis-D. Without the source prior, major limbs are in good alignment, but small body parts are distorted.

%maybe vis net performanc on MPI or MSCOCO, with some pictures. put in supplementary is better.  
\tblallcond    
  
\figAllC
\subsection{Extending Inputs Beyond RGB}
HW-HuP is designed to work well with modalities beyond the commonly-used RGB, which can be helpful under challenging conditions, such as heavy occlusion and total darkness as previously considered in our work \citep{liu2020simultaneously,liu2019seeing}. In this study, we train HW-HuP on individual non-RGB modalities, by replacing the three RGB channels throughout the network with a single channel representing the new modality. The input modalities we work with are long wavelength infrared (LWIR) data, pressure map (PM) data, and the depth data $D$ itself. We also test the combined three-channel modality. Prediction performances under different cover conditions and modalities in \tabref{allcond} show that HW-HuP is still effective under these challenging conditions. As expected, performance is best without the cover, but still reasonable in the other settings. Since the depth is also used to determine the error, it is not surprising that the depth modality results are the strongest. A qualitative result example is shown in \figref{allC}, and more comprehensive results are presented in the \textit{Supplementary Materials}.

\tblHumanD
% 2D projectionr result 

%\tblAllC
% \tblca
%\tblcb
% 4 same table, only report the all cover conditions.  individual cover case show in  supplementary. 

\fighm

\subsection{Large-Scale 3D Human Pose Benchmarks} 
To quantify HW-HuP performance in a setting with 3D ground truth data, we test it on Human3.6M combined with MPII 2D data \citep{andriluka20142d} and compared its mean per joint position error (MPJPE) after Procrustes analysis (PA) alignment with reported results from SOTA approaches in \tabref{humanD}. 
When images without 3D annotation is not available, HW-HuP performs noticeably better over the SOTA. In settings where 3D annotation is available, we note that HW-HuP nonetheless performs better than many SOTA models despite their use of the additional 3D data, and is not too far behind the top 3D enabled models, SPIN \citep{kolotouros2019learning} and  I2L \citep{moon2020i2l}, achieving only 50.4 mm error compared to their 41.1 mm. 
%When 3D annotation is available, while HW-HuP is (as expected) not as good as the top performers, we note that, (1) the gap is relatively small, with HW-HuP 50.4 mm against top performers (SPIN \citep{kolotouros2019learning}, I2L \citep{moon2020i2l}) 41.1mm; (2) HW-HuP still performs better than some SOTAs despite that they have full access to the 3D ground truth. 
To be clear, by design, HW-HuP does not make use of the 3D annotations---the main purpose of this comparison is to quantify the bias of HW-HuP predictions when 3D pose data is not available. 
Some HW-HuP pose predictions on Human3.6M are exhibited in \figref{hm}. The predicted models have good 2D projected alignment on the RGB images, and also match ground truth point clouds quite closely, despite being produced without any ground truth 3D keypoint supervision.

% HW-HuP is initialized with the pre-trained SPIN model, which is trained on data from daily human activities. Nonetheless, 
We also test this model's performance  on the 3D human pose in the wild (3DPW) benchmark \citep{vonMarcard2018}, which focuses on natural outdoor settings. The results in \tabref{DPW} show HW-HuP attaining comparable performance with SOTA approaches despite their full access to the 3D pose ground truth during training.

\tblDPW

%The Human3.6M test is mainly to illustrate the bias when our model converge when we use no gt 3D pose for supervision. 
% Additional results are presented in \textit{Supplementary Materials} including: 
% (1) 3DPW result for 3D human pose in the wild, 
% (2) A real application for 3D infant pose estimation which has not been solved as far as we know. 

\subsection{Infant 3D Pose Estimation}
The analysis of infant motor activity plays a key role in the study of early childhood development \citep{prechtl1990qualitative,hadders1997assessment,zwaigenbaum2013early,hesse2018computer}, and accurate 3D infant pose estimation could enable automation of such analysis. We employ HW-HuP for this challenge, substituting the adult SMPL body model for the infant SMIL model \citep{hesse2018learning}, and testing the result on the MIMM and SyRIP datasets. Note that although statistical infant models are proposed in \citep{hesse2019learning,hesse2018learning}, they do not discuss how to estimate infant pose directly from a RGB image. Rather, they work with sequential RGBD input for both training and inference, and also require a specially configured physical setup to ensure that the infant point cloud can be easily segmented. By comparison, our training can take data from any RGBD setup in the wild, and we only require RGB images for subsequent inference. 
HW-HuP shows aligned depth error on MIMM is 42.0mm against the 58.14mm pretrained SPIN* and 46.2mm on fine tuned SPIN \citep{kolotouros2019learning}. Some qualitative results from both MIMM and SyRIP are shown in \figref{infants}, demonstrating that HW-HuP is able to capture the infant 3D poses in clinical environments as well as in the wild. 

\figInfants   

\section{Conclusion}
%\com{you need to also talk about your infant as well as 3DPW results briefly }
We have introduced a transfer learning strategy for 3D human pose estimation in situations where neither ground truth 3D annotations nor learned priors are available. By selectively learning from source priors and easy-to-access observations in the target domain, our HW-HuP model yields robust 3D pose estimation performance even under challenging conditions. 
%As a practical example, we employed HW-HuP on the in-bed  pose estimation problem, where other SOTA approaches failed. We also demonstrated HW-HuP flexibility with other non-RGB imaging modalities for solving 3D human pose estimation problem in heavily occluded scenarios and under total darkness. Moreover, we reported HW-HuP performance on a popular 3D human pose benchmark, Human3.6M, which showed comparable results with the SOTA approaches that have full access to  the 3D ground truth pose data.  
Since HW-HuP relies only on off-the-shelf depth cameras for training, it has the potential to solve a range of 3D human pose problems in real settings where MoCap training data is infeasible. Possible examples include patient monitoring in medical facilities, infant movement analysis in bassinets or cribs, pilot training observation in the cockpits, and driver behavior or gesture recognition inside of cars. 
%These practical problems will be interesting topics we will explore in future. 

\section{Acknowledgements}
This material is based upon work supported by the National Science Foundation under Grant No. 1755695. %We would like to also thanks our collaborator at Early Markers to give us access to the modeling infant motor movement (MIMM) dataset as a testbed to showcase our algorithm on.

%% The file named.bst is a bibliography style file for BibTeX 0.99c
\bibliographystyle{spbasic}      % basic style, author-year citations
\bibliography{ref}   % name your BibTeX data base

\supp

\end{document}